\documentclass{article}

\usepackage{arxiv}

\usepackage[utf8]{inputenc} %
\usepackage[T1]{fontenc}    %
\usepackage{hyperref}       %
\usepackage{url}            %
\usepackage{booktabs}       %
\usepackage{amsfonts}       %
\usepackage{nicefrac}       %
\usepackage{microtype}      %
\usepackage{lipsum}
\usepackage{graphicx}
\graphicspath{ {./images/} }

\usepackage[utf8]{inputenc} %
\usepackage[T1]{fontenc}    %
\usepackage{hyperref}       %
\usepackage{url}            %
\usepackage{booktabs}       %
\usepackage{amsfonts}       %
\usepackage{nicefrac}       %
\usepackage{microtype}      %
\usepackage{xcolor}         %

\usepackage{researchpack}
\usepackage{verbatim}

\usepackage[dvipsnames,table]{xcolor}
\usepackage{xspace}
\usepackage{enumitem}
\usepackage{wrapfig}
\usepackage{caption}
\usepackage{multirow}
\usepackage{pifont}

\usepackage{natbib} %
\usepackage[capitalise,nameinlink]{cleveref} %

\usepackage{mathtools} %
\usepackage{booktabs} %
\usepackage{tikz} %

\definecolor{acoolcolor}{HTML}{815b9b}
\hypersetup{
    colorlinks=true,
    linkcolor=acoolcolor,
    citecolor=acoolcolor,
}

\definecolor{Tango5}{RGB}{3, 101, 117}
\definecolor{Tango1}{RGB}{247,160,114}

\DeclarePairedDelimiterX{\infdivx}[2]{(}{)}{%
  #1\;\delimsize\|\;#2%
}

\newcommand{\CE}{\ensuremath{\mathsf{CE}}\xspace}

\newcommand{\ECER}{\ensuremath{\mathsf{ECCE}(C)}\xspace}
\newcommand{\change}[1]{#1}
\newcommand{\cchange}[1]{#1}
\newcommand{\acronym}{{Vision-plus-Human-guided CBM}\xspace}
\newcommand{\method}{\texttt{VH-CBM}\xspace}

\newcommand{\CBM}{\texttt{CBM}\xspace}

\newcommand{\CLIP}{\texttt{CLIP}\xspace}
\newcommand{\DINO}{\texttt{DINO}\xspace}

\newcommand{\LABO}{\texttt{LaBo}\xspace}
\newcommand{\LFCBM}{\texttt{LF-CBM}\xspace}
\newcommand{\VLGCBM}{\texttt{VLG-CBM}\xspace}

\newcommand{\SHAPES}{\texttt{Shapes3d}\xspace}
\newcommand{\CelebA}{\texttt{CelebA}\xspace}
\newcommand{\CUB}{\texttt{CUB}\xspace}
\newcommand{\DERMA}{\texttt{Derma}\xspace}

\newcommand{\FY}{\ensuremath{F_1(Y)}\xspace}

\newcommand{\FC}{\ensuremath{F_1(C)}\xspace}

\newcommand{\ROCAUC}{\ensuremath{\mathsf{ROC_{auc}}(C)}\xspace}

\newcommand{\Disent}{\ensuremath{\mathsf{DIS}}\xspace}

\title{Concepts Worth Having: Refining VLM-Guided Concept Bottleneck Models with Minimal Annotations}

\author{
 Nicola Debole\thanks{Corresponding author.} \\
  DISI, University of Trento, Italy\\
  \texttt{nicola.debole@unitn.it} \\
   \And
 Andrea Passerini \\
  DISI, University of Trento, Italy\\
  \texttt{andrea.passerini@unitn.it} \\
  \And
 Stefano Teso \\
  CIMeC, University of Trento, Italy\\
  DISI, University of Trento, Italy\\
  \texttt{stefano.teso@unitn.it} \\
  \And
 Andrea Pugnana\footnotemark[2] \\
  DISI, University of Trento, Italy\\
  \texttt{andrea.pugnana@unitn.it} \\
  \And
 Emanuele Marconato\hspace{0.2em} \thanks{Shared last author.} \\
  DISI, University of Trento, Italy\\
  \texttt{emanuele.marconato@unitn.it} \\
}

\begin{document}
\maketitle
\begin{abstract}
    Concept-bottleneck models (CBMs) are neural classifiers that compute predictions from high-level concepts extracted from the input.  CBMs ensure stakeholders can understand the concepts -- and the predictions they entail -- by learning these from concept-level annotations, which are however seldom available.
    Recent CBM architectures work around this issue by obtaining annotations from Vision-Language Models (VLMs).  While greatly broadening applicability, doing so can yield lower quality concepts and therefore less interpretable models.
    We strike for a middle ground by introducing \acronym (\method), a \textit{hybrid} approach that exploits both VLMs and a small amount of dense annotations.
    \method employs a Gaussian Process in the VLM's embedding space, which captures useful global information about the target domain, to propagate the expert's supervision to any target data point.
    Our empirical evaluation shows how \method predicts more accurate concepts than VLM-guided CBMs even when annotating as little as $1\%$ of the data, \change{while sporting better concept calibration and supporting active learning}.
\end{abstract}

\section{Introduction}

Concept-bottleneck models (CBMs) \citep{koh2020concept} are explainable-by-design neural classifiers that infer predictions from high-level concepts -- \eg objects present in an input image -- they extract from the input. CBMs achieve task performance comparable to black-box alternatives, all while enabling stakeholders to easily \textit{inspect}
the concepts responsible for any prediction.
To support these operations, CBMs encourage the concepts to be interpretable by learning them from \textit{expert annotations}. Yet, these are often unavailable and expensive to obtain, hindering applicability.

State-of-the-art \textit{vision-language-guided CBMs} (henceforth, VLM-CBMs) work around this limitation by replacing expert annotations with weak supervision obtained from large vision-language models (VLMs) \citep{srivastava2024vlgcbm, oikarinen2023label, yang2023language, kazmierczak2024clipqdaexplainableconceptbottleneck}.  While architectural details differ (cf. \cref{sec:preliminaries}), these models enjoy broader applicability while sporting similar or better task performance than CBMs.
The downside is that VLM annotations are not entirely reliable \citep{huang2023survey, sahu2022unpacking, calanzone2025logically}, meaning that VLM-CBMs' concepts tend to be \textit{less accurate} and therefore \textit{less interpretable} \citep{debole2025if}.  This is problematic, as it compromises the promise of interpretability that CBMs are meant to uphold.

\begin{figure*}[!t]
    \centering
    \includegraphics[width=0.95\textwidth]{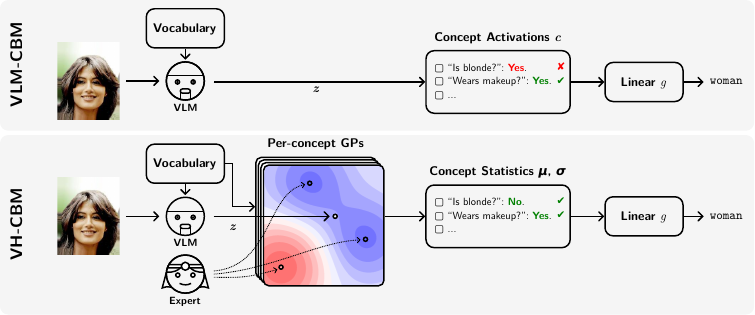}
    \caption{%
    \textbf{Top}: VLM-CBMs rely solely on concepts provided by VLMs -- either at inference time (as above) or for distilling a concept extractor $g$, cf. \cref{sec:preliminaries} -- which can be inaccurate, compromising interpretability \citep{debole2025if}.
    \textbf{Bottom}: \method exploits the VLM's embedding space to propagate expert concept supervision to obtain concept statistics (i.e., $\vmu$ and $\vsigma$ from  the GP) for any test point, improving concept accuracy with few annotations. 
    }
    \label{fig:architecture}
\end{figure*}

We tackle this \change{overlooked}
trade-off by introducing \acronym{s} (\method{s} for short), a simple \textit{hybrid} model that integrates widely available, but unreliable, VLMs with a modicum %
of trustworthy expert supervision.
\method{s} build on the intuition that the VLM embedding space, despite yielding potentially poor \textit{point-wise} annotations, \change{encodes useful} \textit{global} information about the application domain.
\method{s} realize this insight by introducing \change{efficient, stochastic} Gaussian Process\change{es} (GP\change{s}) \citep{williams2006gaussian} in embedding space:  given expert concept annotations for selected examples, the GP\change{s} exploit \change{its} geometry to infer \change{more accurate and calibrated} concept activations for any incoming input.
These are fed to a transparent inference layer to obtain a prediction, cf. \cref{fig:architecture}.

We experimentally evaluate \method{s} on four synthetic and real-world datasets.
Our results indicate that, compared to existing VLM-CBMs, \method{s} predict concepts more accurately from minimal amounts of expert annotations:  even when annotating 
\change{a small portion} of the dataset, \method{s} substantially outperform VLM-CBMs according to different concept-prediction quality metrics, \change{while sporting better concept-level calibration}.
\change{Moreover, increasing the amount of supervision}
monotonically improves concept accuracy, indicating how \method{s} can successfully make use of expert annotations when available, \cchange{especially when allowed to actively query concept annotations}.

\textbf{Contributions}:  Summarizing, we
    (\textit{i}) identify an interpretability-applicability trade-off \change{affecting current} CBM and VLM-CBM architectures;
    (\textit{ii}) introduce \method, a simple approach that addresses this trade-off by mediating between the VLMs and the CBMs with a suitably designed Gaussian Process; and
    (\textit{iii}) demonstrate \change{the promise of \method in terms of concept accuracy and calibration} on several data sets \change{and two backbones} w.r.t. regular \change{and VLM-based} CBMs.

\section{Background}
\label{sec:preliminaries}

\textbf{Notation}.  We abbreviate $\{1, \ldots, n\}$ to $[n]$.
We denote inputs as $\vx \in \bbR^d$ and labels as $y \in [\ell]$.
We consider $k$ (binary or categorical) concepts, each with $v_i \geq 2$ possible values, capturing task-relevant information present in the input, \eg objects and their properties.  We encode them as concept activations in multi-hot format, $\vc \in \bbR^{v}$ with $v :=\sum_{i=1}^k v_i$.
We indicate with $\calX \subseteq \bbR^d$ the input space; with $\calC \subseteq \bbR^v$, $v < d$, the space of concept activations; and with $\calY = [\ell]$ the set of labels.

\textbf{Concept-based Models}.  CBMs sport a two stage architecture comprising (\textit{i}) a \textit{concept extractor} $g: \mathcal{X} \to \mathcal{C}$ that maps the inputs (\eg images of fruits) to concept activations (\eg visual properties such as ``\texttt{red}'' and ``\texttt{round}''), and (\textit{ii}) a \textit{task predictor} $f: \mathcal{C} \to \mathcal{Y}$ -- typically a linear classifier -- that maps concepts to classes (\eg ``\texttt{apple}'' vs. ``\texttt{pear}'').
Given a training set ~$\calD = \{ (\vx^{(i)}, \vc^{(i)}, y^{(i)})\}$ with $n$ examples, %
CBMs can be trained (a) \textit{independently}, \ie $g$ and $f$ are trained separately and later combined; (b) \textit{sequentially}, \ie $g$ is trained first, and its output is used to train $f$; or (c) \textit{jointly}, \ie $g$ and $f$ are trained together by optimizing a joint objective such as
\[
    \sum_{i=1}^n \CE(f(g(\vx^{(i)})), y^{(i)}) + \frac{\lambda}{k} \sum_{j = 1}^k \CE(\hat{c}^{(i)}_j, c^{(i)}_j)
    \label{eq:cbm-joint-ce},
\]
where $\CE$ is the (multi-class) cross-entropy loss and $\hat{c}^{(i)}_j$ is the probability or logit returned by the concept extractor $g$ for the $j$-th concept of $\vx^{(i)}$.
Thanks to this two-stage architecture, CBMs can be \textit{interpretable}, as one can assess which concepts affect the task prediction, and \textit{intervenable}, as users can correct mispredicted concepts to improve the model's predictions~\citep{DBLP:conf/icml/ShinJAL23, DCBMs}.

\textbf{\change{Learning aligned concepts}.}  The interpretability of CBMs
hinges on whether the \change{concepts they learn} are aligned with human semantics \citep{scholkopf2021toward, koh2020concept, marconato2023interpretability}.
Unless this is the case, one can encounter or construct inputs $\vx$ (\eg \change{images of red apples}) on which the CBM and stakeholders infer different concepts (\eg ``orange'' vs. ``red'').  \change{Such} semantic mismatch\change{es} complicate interpretation:  CBM explanations (\eg ``this is an apple because it is \change{red}'') rely on the learned concepts, yet these may not possess the meaning that humans associate to them.
Ensuring that CBMs satisfy this basic property requires training them with an abundance of high-quality concept annotations \citep{koh2020concept, marconato2022glancenets, zarlenga2022concept}.  This is essential, as CBMs trained \textit{without} concept supervision can learn completely misaligned concepts \citep{bortolotti2025shortcuts}.
However, this \change{restricts usage of} CBMs to the few tasks for which high-quality concept supervision is available.

\textbf{Vision-language-guided CBMs}.  Recent CBM architectures work around this issue by leveraging a Vision-Language Model (VLM), such as \CLIP \citep{radford2021learning} or Grounding Dino \citep{liu2024grounding}, to generate weak concept supervision in a zero-shot fashion.
At a high level, VLM-CBMs can be broken down into two steps.
First, if a \textit{concept vocabulary} is not available, they generate one using a foundation model, such as Chat-GPT \citep{achiam2023gpt}.  \change{This amounts to querying}
the \change{latter} for textual descriptions of
concepts that are both \change{discriminative and human understandable},
\eg ``the object is orange'' or ``\change{the object is round}''.  If a concept vocabulary is provided externally, this is used instead.
Next, \change{VLM-CBMs} obtain \textit{weak supervision} for all concepts in the vocabulary.  \change{The specific setup depends on the architecture.}
Some \change{models do so by computing} image-text similarity scores by comparing the \change{embeddings of the} input image and the textual descriptions of the concepts \citep{oikarinen2023label, yang2023language};  others query the VLM directly by asking questions like ``\change{is it true that the object is orange?}'' \citep{srivastava2024vlgcbm};  \change{others still leverage a sparse autoencoder \citep{rao2024discover}}.
The \change{resulting concept activations are either employed} as weak supervision to train a regular CBM, \change{or extracted at inference time and fed to a linear inference layer directly, without the need for training a neural backbone $g$ \citep{yuksekgonul2023post}.}
Our approach follows the latter strategy.

\textbf{Interpretability of VLM-CBMs}.  The main strength of VLM-CBMs is that they enjoy vastly wider applicability compared to regular CBMs, while offering comparable or better task performance.
However, this comes at a cost:  the quality of the concepts they learn strongly depends on the VLM's responses, which can be erratic.  It is well known that VLMs are not entirely reliable, \change{in that} they suffer from hallucinations \citep{huang2023survey} and factual and logical inconsistencies \citep{sahu2022unpacking, calanzone2025logically}.
\change{Such overreliance on VLM annotations implies that VLM-CBMs can learn misaligned concepts:  as shown by \citet{debole2025if}, VLM-CBMs' concepts can be substantially worse than those acquired by regular CBMs according to several quality metrics, including concept accuracy, undermining their interpretability.}

\section{Hybrid Human-and-Vision Guided CBMs}
\label{sec:method}

In order to meet its promises, any CBM should satisfy at least the following desiderata:
\begin{itemize}[leftmargin=2em]

    \item[\textbf{D1}] \textbf{\textit{Task accuracy}}: it should output high-quality task predictions.

    \item[\textbf{D2}] \textbf{\textit{Concept accuracy}}: it should learn accurate concepts, so as to support interpretation.

    \item[\textbf{D3}] \textbf{\textit{Applicability}}: it should rely on a minimal amount of expert annotations.

\end{itemize}
Importantly, \textit{existing architectures fail at least one of our desiderata}. Regular CBMs are not widely applicable, failing \textbf{D3}, while VLM-CBMs suffer from impaired \change{concept accuracy}, thus failing \textbf{D2}.
\change{This suggests current CBM architectures face a}
trade-off between interpretability and applicability:  in order to learn highly accurate concepts, it is paramount that CBMs have access to high-quality concept labels, and these are typically only obtainable from experts.  Naturally, this requirement restricts applicability.  At the same time, completely relying on weak labels produced by VLMs, while improving applicability, can drastically reduce concept accuracy \citep{debole2025if}, and thus interpretability.

\textbf{Beyond concept accuracy}.  Desideratum \change{\textbf{D2} focuses on concept accuracy, as this is a hard prerequisite for interpretability:  unless a human and a CBM infer the same concepts for the same inputs, they might not agree on what those concepts mean.
However, concept quality encompasses other key factors.
An important metric is disentanglement~\citep{suter2019robustly,eastwood2018framework, higgins2018towards,scholkopf2021toward,marconato2022glancenets,debole2025if}, which measures to what extent concepts encode unwanted information about each other.  While higher accuracy can benefit disentanglement \citep{marconato2022glancenets}, this is not always the case.  At the same time, concepts should be calibrated, so that users can reliably infer whether the learned concepts and their semantics can be trusted \citep{marconato2024bears}.
Hence, in our experiments we will evaluate accuracy, calibration, and disentanglement.}

\subsection{The \method Framework}
\label{sec:method}

We tackle \change{the above} trade-off by introducing \method, a \change{simple but effective} approach combining the strengths of regular and vision-guided CBMs.  \method produces highly accurate task (\textbf{D1}) and concept (\textbf{D2}) predictions by integrating two sources of information:  the VLM's embedding space and a modicum of expert supervision.  The latter is kept to a minimum so as to retain applicability (\textbf{D3}).

\textbf{Overview}.  \change{A \method comprises three components, cf.} \cref{fig:architecture}:
\textit{i}) a VLM, mapping the inputs to a well-structured embedding space;
\textit{ii}) \change{one} \textit{Gaussian Process} (GP) classifier \citep{williams2006gaussian} \change{for each concept (\eg one for shape and one for color), living in the VLM's embedding space}.  \change{These are} responsible for estimating \change{the mean and variance of the concept activations}; \change{binary concepts (\eg wearing makeup) are predicted with binary GPs and categorical concepts (\eg color) with multi-class GPs}; and
\textit{iii}) a linear inference layer that receives the per-concept statistics and infers a prediction.
Essentially, the GP\change{s} act as a middle man between the VLM and the inference layer, propagating the expert annotations by following the geometry of the embedding space, \change{while providing all benefits of Bayesian models, such as improved calibration}.

\textbf{Training and inference}.  \change{Like other CBMs, \method assumes access to a concept vocabulary -- defined either manually or through a foundation model -- to determine the set of target concepts}.  \change{It also requires} two data sets:  a larger data set $\calD = \{ (\vx^{(i)}, y^{(i)} : i \in [n] \}$ annotated with task labels only, used \change{for} training, and a smaller subset $\calD_s \subset \calD$ of $n_s \ll n$ examples accompanied by expert annotations $\vc^{(i)}$, used also at inference time. %
\change{\method are trained as follows}.
    First, one obtains the VLM embeddings $\vz^{(i)} \in \bbR^d$ for all inputs $\vx^{(i)} \in \calD$, \change{which are then standardized
    and normalized.}
    Next, \change{each GP classifier
    is fit on the annotated data $\calD_s$, as explained below.}
    Finally, \change{the GPs' predictions are employed} to train a linear layer on the entire training set $\calD$, thus implementing a form of \textit{sequential} training \change{(see \cref{sec:preliminaries})}.

For any new input $\vx$, \change{the} \method\ \change{sequentially applies the VLM, GPs and inference layer}
to obtain the concept activations and a prediction.
\change{Crucially, \textit{the GPs' predictions exploit the expert's supervision $\calD_s$, propagating it according to the VLM's embedding space}.}
Next, we discuss the GP and inference layer in detail.

\textbf{Inferring Concepts with Gaussian Processes}.  GPs are a well-known class of Bayesian kernel-based models \citep{williams2006gaussian, nickisch2008approximations}.  \method employs them for predicting concept activations from given densely annotated examples, working directly on the embeddings $\vz$.
Note that \method has to predict multiple concepts (\eg color and shape) for each example, and that each of them can be binary or categorical.

\change{To this end}, we adapt the multi-class architecture of \citet{teh2005semiparametric}, stacking together \change{one GP for each concept} to predict all concepts jointly.
\change{To illustrate how this works},
we focus on a single categorical concept $c_i$ for some $i \in [k]$, say, color, \change{with $v_i \geq 2$ possible values.
We want to compute an activation for each such value while taking into account their mutual exclusivity: color can be either red or blue, but not both. Taking these dependencies into account helps performance -- as confident positive predictions suppress all alternatives, most of which are incorrect -- and reduces sample complexity.}
\change{We follow the same approach as \citet{milios2018dirichlet} and perform regression on Dirichlet-transformed labels instead of solving a classification task. This Dirichlet formulation allows us to produce well-calibrated probability estimates, avoiding the intractable inference required by standard multi-class GP classifiers.}
Moreover, similarly to \citet{teh2005semiparametric}, we introduce $v_i$ \textit{latent} GP regressors, indexed by $j \in [v_i]$.  Each GP is parameterized by a mean function $m_{ij}: \bbR^d \to \bbR$ over the embeddings, which acts as a prior on the value's score, and a covariance function (or kernel) $K_{ij}: \bbR^d \times \bbR^d \to \bbR$ that defines a similarity between embeddings. Together, they determine a distribution over real-valued functions 
\[
    s_{ij} (\vz)
    \sim 
    \mathcal{GP} \big(m_{ij},
    K_{ij}\mid \vz) \, ,
\]
mapping from embeddings $\vz$ to real-valued scores.  Following \citet{teh2005semiparametric}, we then linearly mix these values together as $\vA_i^\top \vs_i(\vz)$, \change{where the vector $\vs_i$ captures all $j$-th scores of the concept $c_i$, and $\vA_i \in \bbR^{v_i \times v_i}$ denotes a learnable matrix used to correlate the individual components}. 
The \change{resulting} $j$-th component $(\vA_i^\top \vs_i(\vz))_j$
\change{acts as a} logit $\hat{c}$ for concept $i$ attaining value $j$, cf. \cref{eq:cbm-joint-ce}.
Finally, the predictive distribution over color values is computed as the softmax of the transformed scores, in expectation over the distribution induced by the $v_i$ latent GPs, namely
\[
    p(c_i =j \mid \vz)
    = 
    \bbE_{\vs_i(\vz) \sim \calG \calP_i}  
    \left[
    \mathrm{softmax} \big( 
    \vA_i^\top \vs_i(\vz)
    \big)_j
    \right] \, .
\]
Here, $\calG\calP_i = \prod_{j} \calG \calP(m_{ij}, K_{ij} \mid \vz)$ is the joint distribution over all concepts.
In practice, we model the mean function as a learnable constant and employ an RBF kernel, \ie
\[
    \textstyle
    K_{ij}(\vz^{(a)}, \vz^{(b)}) = \alpha_{ij}^2\exp \big( -  \lVert \vz^{(a)} - \vz^{(b)} \lVert^2  / 2\rho_{ij}^2 \big) \, .
\]
The hyper-parameter $\alpha_{ij}$ is the output scale and $\rho_{ij}$ is the length scale which are \change{fit on data}.

The mixing matrix $\vA_i$ is also learned during training, as done by \citet{teh2005semiparametric}. To determine the prior on the concept score $m_{ij}$, selecting inducing points of the kernels $K_{ij}$ and their hyper-parameters, we employ a stochastic variant of GPs \citep{hensman2013gaussian} that maximises the Evidence Lower Bound (ELBO). The ELBO is computed between the latent GP posterior $\calG \calP_i$, and the Dirichlet-transformed \citep{milios2018dirichlet} concept annotations, which proved effective in predicting concept activations in representation space \citep{wang2023gaussian}. 
The ELBO objective balances the complexity of the GP model (via the KL divergence from the prior) and the accuracy of the predicted class probabilities $p(c_i \mid \vz)$ relative to the annotated concept value for the embedding $\vz$. 
Additional details can be found in \cref{sec:appendix-impl-details}.

\textbf{Fitting the CBM}. 
We use the $v$ concept activations to fit a linear layer \change{$f$ mapping concepts to labels}.
This takes as input the stacked concept scores $(\vA_i^\top \vs_i(\vz))$ sampled from the GP. 
Let $\vs(\vz) := (\vs_1(\vz)^\top, \ldots, \vs_k(\vz)^\top )^\top \in \bbR^{v}$,
be the overall concept score for all $k$ concepts sampled from the
\change{joint} GP distribution ${\calG \calP} = \prod \calG\calP_i$, and $\vA = \mathrm{diag}(\vA_1, \ldots, \vA_k) \in \bbR^{v \times v}$ be the block-diagonal mixing matrix.  
We train a linear classifier with weights
$\vW = (\vw_1, \ldots, \vw_\ell) \in \bbR^{v \times \ell} $ and bias $\vb \in \bbR^\ell$ to maximize the likelihood of the true label for all pairs $(\vz, y) \in \calD$, \change{namely}:
\[
    \log p(y \mid \vz) =
    \log \, \bbE_{ \vs \sim \calG \calP } \left[
    \mathrm{softmax}
    \big(
        \vW^\top \vA^\top \vs + \vb
    \big)_y
    \right] \, .
\]
\change{As with other CBMs, the prediction can be explained in terms of activated concepts, \eg by presenting stakeholders their mean activations}.

\textbf{Uncertainty and active learning}.  \cchange{As we will see in \cref{sec:experiments}, propagating expert supervision through a Bayesian model helps \method achieve substantially better concept calibration.  This is useful for further reducing annotation cost through uncertainty-based active learning \citep{settles2012active}.  In practice, in our experiments we employ the GP's predictive uncertainty to prioritize what concepts (but not what examples) to annotate.  We provide further details in \cref{sec:active-selection}.}

\subsection{Benefits and Limitations}
\label{sec:benefits-and-limitations}

\textbf{Annotation cost}.  \change{\method{s} aim} at improving concept accuracy by integrating expert annotations \change{into the VLM-CBM pipeline}.
The main downside is the annotation cost.  We argue this is limited and well justified.  Our experiments in \cref{sec:experiments} indicate that \change{few concept annotations} can substantially boost concept accuracy, which is essential, especially in high-stakes applications where interpretability is a hard requirement. \change{Moreover, as we will show in \cref{sec:experiments}, the GP uncertainty estimates improve the efficacy of active learning, further reducing annotation costs.}
Thus, \method{s} sport better applicability (\textbf{D3}) than regular CBMs while performing as well as VLM-CBMs in terms of task accuracy (\textbf{D1}) and much better in terms of concept accuracy (\textbf{D2}).

\textbf{Scalability}. \change{A second concern is scalability.  We remark that, in our case, GP inference scales gracefully with the number of concepts $k$ (specifically, \textit{linearly}) and with annotated set size.  This is achieved by employing inducing points \citep{snelson2005sparse}. Moreover, the per-concept GPs are fit independently, making parallelization possible. We report  further implementation details in~\cref{sec:hyperparameters}. }

\change{In addition, despite having to invoke a VLM at inference time, \method is still efficient in practice: the embedding step relies on a frozen encoder, which requires no gradient computation and can be efficiently batched on modern hardware.  This is in line with other VLM-CBMs \citep{yuksekgonul2023post, rao2024discover, yang2023language}.
However, an alternative is to replace the VLM at inference time with a proper concept extractor distilled from the GP's annotations \citep{oikarinen2023label, lai2023faithful, srivastava2024vlgcbm}, potentially using inverse propensity scoring or distribution matching to avoid transferring information about concepts the GP is unsure about.}

\textbf{Access to embeddings}. Finally, \method{s} require access to the VLM's embedding space. This is by design:  the \textit{global} geometry of the embedding space tends to be more reliable than its \textit{local} weak annotations.  Our results support this observation: in practice, the baselines using only \CLIP's annotations underperform compared to both regular CBMs and \method. \change{This holds for two distinct VLM backbones: \CLIP \citep{radford2021learning}, the de-facto standard for contrastive image-textual embeddings; and \DINO(v3) \citep{simeoni2025dinov3}, a state-of-the-art self-supervised vision backbone.}

\section{Empirical Analysis}
\label{sec:experiments}

We address empirically the following research questions:
\begin{itemize}[leftmargin=2em]

    \item[\textbf{Q1}] 
    Can \method improve concept accuracy with a small set of annotated samples?

    \item[\textbf{Q2}] Does \method{}  \change{improve concept calibration}?  
    
    \item[\textbf{Q3}] Is \method competitive in terms of task accuracy?

\end{itemize}

\begin{wraptable}[9]{r}{7.5cm}
    \centering
    \caption{\change{Data sets statistics.}}
    \scalebox{0.8}{
    \begin{tabular}{cccc}
    \toprule
    {\sc Dataset}
        & {\sc Splits}
        & {\sc \# Concepts}
        & {\sc \% Positives}
    \\
        \midrule
        \SHAPES
        & $48$k/$5$k/$30$k %
        & $42$ (Mixed)
        & $14\%$ %
    \\
        \CelebA
        & $25$k/$5$k/$20$k %
        & $39$ (Multi-hot)
        & $22\%$ %
    \\
        \CUB
        & $4.8$k/$1.2$k/$5.8$k %
        & $112$ (Multi-hot)
        & $20\%$ %
    \\
        \DERMA
        & $7$K/$1$K/$2$K %
        & $7$ (One-hot)
        & $12\%$ %
    \\
    \bottomrule
    \end{tabular}
    }
    \label{tab:datasets}
\end{wraptable}
\textbf{Data sets}. We evaluate our approach on four datasets \change{with both concept and task annotations}, \change{cf. \cref{tab:datasets} for an overview}.
{\tt \underline{Sha}p\underline{es3d}}~\citep{kim2018disentangling} consists of
pre-rendered images of 3D objects with distinct shapes, colors, orientations and background, amounting to 5 multiclass concepts corresponding to a total of $42$ binary concept activations. The task is to discriminate images containing a red pill object. We consider the original data splits, i.e., $48$k samples for training, $5$k for validation and $30$k for evaluation.
\underline{\CelebA} \citep{liu2015deep} includes $200$k images of celebrity faces described through $40$ binary concepts (\eg hair color and presence of makeup).  The task is to predict the gender from the other concepts.  \change{Following \citet{debole2025if}, we randomly sampled $25$k examples for training, $5$k for validation, and $20$k for evaluation}.
\underline{\CUB} \citep{wah2011caltech} consists of bird images described by 112 concepts (\eg bill shape, breast color) corresponding to $112$ binary activations. The task is to distinguish among $200$ possible bird species. We consider the original data splits, \ie \change{$4.8$k samples for training, $1.2$k for validation and $5.8$k for evaluation}.
\underline{\DERMA}~\citep{medmnistv1, medmnistv2} is a medical dataset containing about $10$k images of skin lesions; each sample is annotated from an expert with one of $7$ lesion types that range from malign to benign. The task is to predict lesion is malign or benign. \change{We include this dataset as a challenging case where \CLIP and \DINO embeddings are suboptimal, since visual encodings of skin lesion images yield poorly discriminative concept representations.} We use the original data splits with a $70/10/20\%$ proportion for training, validation and testing, respectively.

\textbf{Methods}. For all the research questions, we compare \method{} \change{(leveraging either \CLIP or \DINO embeddings)} against a \change{CBM with full concept supervision} (\underline{\texttt{CBM}@$100$}) and three state-of-the-art VLM-CBMs.
Specifically, \underline{\LABO} \citep{yang2023language} uses the cosine similarity between image embeddings and textual concept embeddings of \CLIP as a score for each concept, which is then used to predict task labels. %
\underline{\LFCBM} \citep{oikarinen2023label} also uses similarity between image and text embeddings from \CLIP, but it distills a concept encoder (with a different backbone) to match those similarities.
Finally, \underline{\VLGCBM} \citep{srivastava2024vlgcbm} leverages Grounding DINO \citep{liu2024grounding} to annotate each sample with both concepts and bounding-boxes (accompanied by confidence scores).
Additional implementation details can be found in \cref{sec:appendix-impl-details}.

For \textbf{Q1}, we compare \method against two additional baselines \change{trained using a randomly drawn percentage of concepts annotations (reported after the ``@'' symbol):
(\textit{i}) a CBM trained sequentially from scratch (\texttt{CBM}-@)~\citep{koh2020concept},
and (\textit{ii}) a linear probe~(\texttt{LP}-@) trained to map from CLIP embeddings to concepts \citep{alain2016understanding}.
}
We also consider a linear probe trained using \textit{all} available concept annotations (\texttt{LP-All}) and, as a lower bound, the results obtained using CLIP scores in a zero-shot fashion (\texttt{CLIP}).

\begin{figure*}[t]
    \centering
    \includegraphics[width=\linewidth]{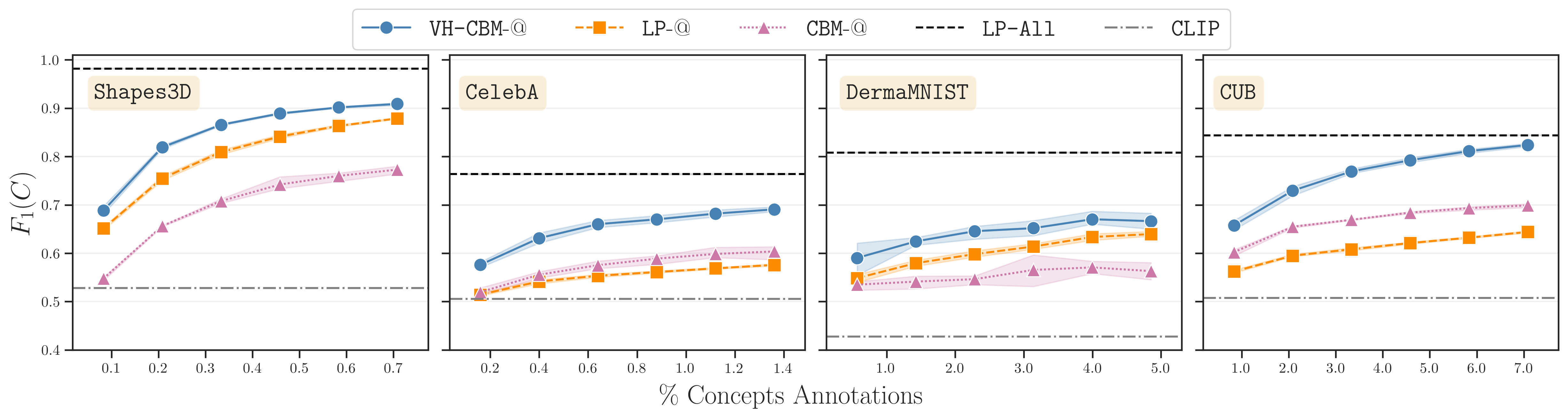}

    \includegraphics[width=\linewidth]{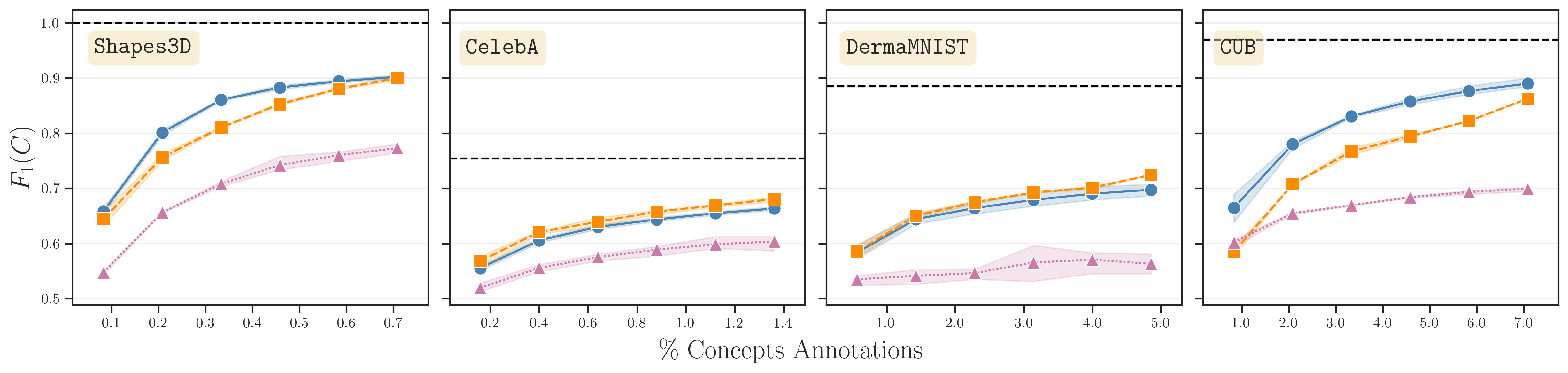}
    \caption{\textbf{\change{\method improves concept accuracy.}}
    \FC of \method with varying percentages of concept supervision for the \CLIP (top) and \DINO (bottom) backbones.  We report average \FC over three runs, with confidence intervals estimated via bootstrap resampling across runs.
    }
    \label{fig:fig2-clip}
    \label{fig:fig2-dino}
\end{figure*}

\textbf{Metrics}. For all competitors, we assess how close they come to satisfying \textbf{D1} and \textbf{D2} by evaluating task predictions and \change{concept accuracy}, respectively.
For \textbf{Q1}, we evaluate concept predictions using per-concept macro $F_1$ score, and report the average across the $k$ concepts \FC. In addition, we track the Area under the Receiver operating characteristic Curve of the concepts, denoted \ROCAUC, \change{and also disentanglement (\Disent) among concepts using the DCI metric \citep{eastwood2018framework,eastwood2022dci}}. This gives an indication about how intertwined unrelated concepts (like shape and color)  are (see \cref{sec:method} for a brief discussion).
We refer the reader to \cref{sec:metrics-appx} for \change{details about the metric}.
For \textbf{Q2}, we evaluate the \change{empirical cumulative calibration error} (\ECER) \citep{arrieta2022metrics}, as it does not depend on binning choices. We provide additional calibration results and details in \cref{tab:full-calibration-results} and in \cref{sec:calibration-appendix}. %
For \textbf{Q3}, we measure task performance using macro $F_1$ score, denoted \FY, to avoid imbalancedness concerns.

\textbf{Setup.} For each dataset, we $(i)$ start with 40 randomly fully annotated samples; $(ii)$ vary the amount of concept supervision by adding 60 newly concept-annotated \change{samples} per iteration, for a total of five iterations; \change{and} $(iii)$ at each supervision level, train each method on the available annotated samples, and evaluate performance on the test set.
\cchange{In step $(ii)$, for \method we exploit the GP's uncertainty to select the most uncertain \textit{concepts} for annotation, while for the competitors we annotate all concepts; full results for the non-active variant of \method are left to \cref{sec:random_results_appendix}.}
\cchange{In either case,} \change{at the final iteration (\ie the last point of the curve in \cref{fig:fig2-clip}, from which the results in \cref{tab:results} are computed), 340 concepts have been annotated. Since dataset sizes vary, this corresponds to different annotation percentages across datasets (e.g., in \SHAPES, which has a training set of 48{,}000 samples, the annotated ratio is 0.71\%). For reference, we report these final ratios in \cref{tab:results}; further details are provided in \cref{sec:pipeline-appendix}.} We repeat this procedure, \change{for both \CLIP and \DINO embeddings,} three times using different seeds.

\begin{table*}[!t]
    \centering
    \caption{\textbf{\method{@} is more accurate \change{and calibrated}} \change{compared to other VLM-CBMs and \CBM{@} across datasets when trained on $340$ concept annotated samples}.  Best is in \textbf{bold}, second-best \underline{underlined}. \CBM{@100\%} is a fully supervised baseline so it is excluded from the ranking.
    }
    \scalebox{0.85}{
    \begin{tabular}{llccccc}
        \toprule
        & {\sc Model} & \FY ($\uparrow$) & \FC ($\uparrow$) & \ROCAUC ($\uparrow$) & \Disent ($\uparrow$) & \ECER ($\downarrow$) \\
        \cmidrule{1-7}
        \multirow[c]{7}{*}{\rotatebox{90}{\SHAPES}}
        & \CBM @ $100\%$
            & $0.99 \pm 0.01$     %
            & $0.99 \pm 0.01$     %
            & $0.99 \pm 0.01$ %
            & $0.99 \pm 0.01$   
            & $0.38 \pm 0.07$   \\ %
                    \cmidrule{2-7}
        & \cellcolor{gray!10}\CBM @  $0.71\%$
            & \cellcolor{gray!10}$0.94 \pm 0.01$     %
            & \cellcolor{gray!10}$0.77 \pm 0.01$     %
            & \cellcolor{gray!10}$0.92 \pm 0.01$ %
            & \cellcolor{gray!10}$0.63 \pm 0.01$   
            & \cellcolor{gray!10}$0.39 \pm 0.07$   \\ %

        & \LABO
            & $0.76 \pm 0.01$     %
            & $0.56 \pm 0.01$     %
            & $0.72 \pm 0.01$ %
            & $0.28 \pm 0.01$   
            & $0.89 \pm 0.02$   \\ %
        & \cellcolor{gray!10}\LFCBM
            & \cellcolor{gray!10}$0.96 \pm 0.01$     %
            & \cellcolor{gray!10}$0.57 \pm 0.01$     %
            & \cellcolor{gray!10}$0.72 \pm 0.01$ %
            & \cellcolor{gray!10}$0.28 \pm 0.01$   
            & \cellcolor{gray!10}$0.39 \pm 0.29$   \\ %
        & \VLGCBM
            & $0.49 \pm 0.06$     %
            & $0.54 \pm 0.01$     %
            & $0.68 \pm 0.01$ %
            & $0.24 \pm 0.01$   
            & $0.31 \pm 0.25$   \\ %
        
        & \cellcolor{blue!10}\method \texttt{CLIP} @ $0.71\% $
            & \cellcolor{blue!10}$\mathbf{0.99 \pm 0.01}$     %
            & \cellcolor{blue!10}$\mathbf{0.91 \pm 0.01}$     %
            & \cellcolor{blue!10}$\underline{0.95 \pm 0.01}$ %
            & \cellcolor{blue!10}$\mathbf{0.81 \pm 0.01}$   
            & \cellcolor{blue!10}$\mathbf{0.03 \pm 0.01}$   \\ %
        
        & \cellcolor{blue!10}\method \texttt{DINO} @ $0.71\% $
            & \cellcolor{blue!10}$\underline{0.98 \pm 0.01}$     %
            & \cellcolor{blue!10}$\underline{0.89 \pm 0.01}$     %
            & \cellcolor{blue!10}$\mathbf{0.96 \pm 0.01}$ %
            & \cellcolor{blue!10}$\underline{0.79 \pm 0.03}$   
            & \cellcolor{blue!10}$\mathbf{0.03 \pm 0.01}$   \\ %
            
        \midrule\midrule

        \multirow[c]{7}{*}{\rotatebox{90}{\CelebA}}
        & \CBM @ $100\%$
            & $0.96 \pm 0.01$     %
            & $0.77 \pm 0.01$     %
            & $0.93 \pm 0.01$ %
            & $0.75 \pm 0.01$   
            & $0.32 \pm 0.15$   \\ %
                    \cmidrule{2-7}
        & \cellcolor{gray!10}\CBM @ $1.4\%$
            & \cellcolor{gray!10}$0.90 \pm 0.01$     %
            & \cellcolor{gray!10}$0.60 \pm 0.01$     %
            & \cellcolor{gray!10}$0.81 \pm 0.01$ %
            & \cellcolor{gray!10}$0.37 \pm 0.02$   
            & \cellcolor{gray!10}$0.31 \pm 0.14$   \\ %
        & \LABO
            & $\underline{0.98 \pm 0.01}$     %
            & $0.55 \pm 0.01$     %
            & $0.66 \pm 0.01$ %
            & $0.28 \pm 0.01$   
            & $0.77 \pm 0.20$   \\ %
        & \cellcolor{gray!10}\LFCBM
            & \cellcolor{gray!10}$\mathbf{0.99 \pm 0.01}$     %
            & \cellcolor{gray!10}$0.58 \pm 0.01$     %
            & \cellcolor{gray!10}$0.68 \pm 0.01$ %
            & \cellcolor{gray!10}$0.29 \pm 0.01$   
            & \cellcolor{gray!10}$0.40 \pm 0.24$   \\ %
        & \VLGCBM
            & $\underline{0.98 \pm 0.01}$     %
            & $0.54 \pm 0.01$     %
            & $0.62 \pm 0.01$ %
            & $0.22 \pm 0.01$   
            & $0.42 \pm 0.27$   \\ %
            
        & \cellcolor{blue!10}\method \texttt{CLIP} @ $1.4\% $
            &\cellcolor{blue!10}$\underline{0.98 \pm 0.01}$     %
            &\cellcolor{blue!10}$\mathbf{0.69 \pm 0.01}$     %
            & \cellcolor{blue!10}$\underline{0.82 \pm 0.01}$ %
            & \cellcolor{blue!10}$\underline{0.40 \pm 0.01}$   
            & \cellcolor{blue!10}$\mathbf{0.04 \pm 0.03}$   \\ %
                
        & \cellcolor{blue!10}\method \texttt{DINO} @ $1.4\% $
            & \cellcolor{blue!10}$0.97 \pm 0.01$     %
            & \cellcolor{blue!10}$\underline{0.65 \pm 0.01}$     %
            & \cellcolor{blue!10}$\mathbf{0.83 \pm 0.01}$ %
            & \cellcolor{blue!10}$\mathbf{0.43 \pm 0.01}$   
            & \cellcolor{blue!10}$\mathbf{0.04 \pm 0.03}$   \\ %

        \midrule\midrule %

        \multirow[c]{7}{*}{\rotatebox{90}{\DERMA}}
        & \CBM @ $100\%$
            & $0.68 \pm 0.03$     %
            & $0.65 \pm 0.01$     %
            & $0.91 \pm 0.00$ %
            & $0.28 \pm 0.03$   
            & $0.41 \pm 0.11$   \\ %
                    \cmidrule{2-7}
        & \cellcolor{gray!10}\CBM @ $4.8\%$
           & \cellcolor{gray!10}${0.65 \pm 0.05}$     %
            & \cellcolor{gray!10}$0.56 \pm 0.02$     %
            & \cellcolor{gray!10}$0.80 \pm 0.02$ %
            & \cellcolor{gray!10}$0.15 \pm 0.03$   
            & \cellcolor{gray!10}$0.40 \pm 0.13$   \\ %
        & \LABO
            & ${0.57 \pm 0.01}$     %
            & $0.42 \pm 0.01$     %
            & $0.57 \pm 0.01$ %
            & $0.04 \pm 0.01$   
            & $0.86 \pm 0.22$   \\ %
        & \cellcolor{gray!10}\LFCBM
            & \cellcolor{gray!10}$0.51 \pm 0.03$     %
            & \cellcolor{gray!10}$0.42 \pm 0.01$     %
            & \cellcolor{gray!10}$0.56 \pm 0.01$ %
            & \cellcolor{gray!10}$0.05 \pm 0.01$   
            & \cellcolor{gray!10}$0.14 \pm 0.22$   \\ %
        & \VLGCBM
            & $0.43 \pm 0.01$     %
            & $0.42 \pm 0.01$     %
            & $0.50 \pm 0.01$ %
            & $0.02 \pm 0.01$   
            & $0.27 \pm 0.35$   \\ %

        & \cellcolor{blue!10}\method \texttt{CLIP} @ $4.8\%$
            & \cellcolor{blue!10}$\underline{0.67 \pm 0.01}$     %
            & \cellcolor{blue!10}$\mathbf{0.67 \pm 0.03}$     %
            & \cellcolor{blue!10}$\underline{0.88 \pm 0.02}$ %
            & \cellcolor{blue!10}$\underline{0.28 \pm 0.04}$   
            & \cellcolor{blue!10}$\mathbf{0.02 \pm 0.01}$   \\ %

        & \cellcolor{blue!10}\method \texttt{DINO} @ $4.8\%$
            & \cellcolor{blue!10}$\mathbf{0.69 \pm 0.01}$     %
            & \cellcolor{blue!10}$\underline{0.66 \pm 0.01}$     %
            & \cellcolor{blue!10}$\mathbf{0.92 \pm 0.01}$ %
            & \cellcolor{blue!10}$\mathbf{0.33 \pm 0.01}$   
            & \cellcolor{blue!10}$\mathbf{0.02 \pm 0.01}$   \\ %

        \midrule\midrule

        \multirow[c]{7}{*}{\rotatebox{90}{\CUB}}
        & \CBM @ $100\%$
            & $0.69 \pm 0.01$     %
            & $0.86 \pm 0.01$     %
            & $0.97 \pm 0.01$ %
            & $0.75 \pm 0.01$   
            & $0.32 \pm 0.13$   \\ %
                    \cmidrule{2-7}
        & \cellcolor{gray!10}\CBM @ $7.0\%$
            & \cellcolor{gray!10} $0.63 \pm 0.01$     %
            & \cellcolor{gray!10}$0.70 \pm 0.01$     %
            & \cellcolor{gray!10}$0.84 \pm 0.01$ %
            & \cellcolor{gray!10}$0.37 \pm 0.01$   
            & \cellcolor{gray!10}$0.32 \pm 0.12$   \\ %
        & \LABO
            & $0.27 \pm 0.01$     %
            & $0.56 \pm 0.01$     %
            & $0.66 \pm 0.01$ %
            & $0.19 \pm 0.01$   
            & $0.80 \pm 0.13$   \\ %
        & \cellcolor{gray!10}\LFCBM
            & \cellcolor{gray!10}${0.67 \pm 0.01}$     %
            & \cellcolor{gray!10}$0.48 \pm 0.01$     %
            & \cellcolor{gray!10}$0.45 \pm 0.01$ %
            & \cellcolor{gray!10}$0.19 \pm 0.01$   
            & \cellcolor{gray!10}$0.21 \pm 0.09$   \\ %
        & \VLGCBM
            & $0.60 \pm 0.01$     %
            & $0.79 \pm 0.01$     %
            & $0.91 \pm 0.01$ %
            & $\underline{0.61 \pm 0.01}$   
            & $0.23 \pm 0.13$   \\ %

        & \cellcolor{blue!10}\method \texttt{CLIP} @ $7.0\%$
            & \cellcolor{blue!10}$\underline{0.80 \pm 0.01}$     %
            & \cellcolor{blue!10}$\underline{0.82 \pm 0.01}$     %
            & \cellcolor{blue!10}$\underline{0.92 \pm 0.01}$ %
            & \cellcolor{blue!10}$0.56 \pm 0.01$   
            & \cellcolor{blue!10}$\mathbf{0.02 \pm 0.01}$   \\ %

        & \cellcolor{blue!10}\method \texttt{DINO} @ $7.0\%$
            & \cellcolor{blue!10}$\mathbf{0.88  \pm 0.01}$     %
            & \cellcolor{blue!10}$\mathbf{0.89  \pm 0.01}$     %
            & \cellcolor{blue!10}$\mathbf{0.94  \pm 0.01}$ %
            & \cellcolor{blue!10}$\mathbf{0.71  \pm 0.01}$   
            & \cellcolor{blue!10}$\mathbf{0.02 \pm 0.01}$   \\ %
            
        \bottomrule
    \end{tabular}
    }
    \label{tab:results}
\end{table*}

\textbf{Q1: \method{s} help improve concept accuracy}.  In \cref{fig:fig2-clip}, we report the average \FC for \method and competitors, \change{for both backbones}, when varying the percentage of annotated concepts.
\change{Across datasets, \method improves over the competitors in $6$ cases out of $8$, often with a sizable performance gap.  In the remaining two case (\CelebA and \DERMA on \DINO) it performs on par with {\tt LP-@}.  We speculate this occurs because in these cases the concepts are linearly retrievable, meaning a simple linear model is sufficient.}
The largest gains can be seen on \texttt{CUB}:  with less than 1\% of training concepts annotated, \method improves by $\approx 31.8\%$ with respect to training-free \CLIP scores, and by $\approx 11\%$ when compared to the runner up, \texttt{CBM-}@.
\change{A similar gain also appears with  \DINO as a backbone, see \cref{fig:fig2-clip} (bottom): in a single iteration \method surpasses the competitors.}
The upward trend across all datasets shows how \change{adding more} concept supervision keeps improving \FC. 
\change{With \CLIP this effect is more pronounced,}
\eg  we can see that on \texttt{CUB}, with just 7\% of annotated instances, 
the \FC gap with respect to \texttt{LP-All} is $\approx3.1$ p.p., suggesting that \method provides the best trade-off between required annotations and higher concept accuracy.
\change{On \DINO, improvements are more pronunced in medium regimes, but \texttt{LP-@} fares well (surpassing \method on \CelebA and \DERMA).
}

In \cref{tab:results} the comparison to other VLM-CBMs shows that \method{@} substantially improve \FC and \ROCAUC, with a high margin in all datasets, \change{with the other supervised baseline \CBM{@} lagging behind}.
\change{\VLGCBM is the VLM-based runner up, ranking slight below \method in \CUB, but sensibly worsening \FC and \ROCAUC in the other datasets
(\method improves to \VLGCBM at most by $\approx 37$ p.p. on \FC in \SHAPES and almost $25$ p.p. in \ROCAUC on \DERMA).
}

\textbf{Q2: \method yields more calibrated \change{and disentangled} concepts}. 
\change{
\cref{tab:results} shows that \method consistently achieves near-zero \ECER across all datasets ($\approx 0.02\text{--}0.04$), substantially outperforming all competing approaches, including \change{all VLM-CBMs and the} partially and fully supervised CBMs.
In contrast, \change{these} often exhibit \change{much} higher calibration error, \eg \LABO reaches up to $\approx 0.89$ on \SHAPES. Even when comparing \method to linear probes trained on the same embeddings (reported in \cref{tab:full-calibration-results}), \method shows improved calibration, suggesting that the GP backbone plays a crucial role in producing calibrated uncertainty estimates. %
}

\change{
In \cref{fig:active-vs-rand-clip,fig:active-vs-rand-dino}, we report an ablation comparing random vs. active learning variants, which shows how \method's concepts allow to better identify uncertain or informative regions of the embedding space for both \CLIP and \DINO, making it more likely to query annotations that lead to larger improvements. Nonetheless, the random variant of \method still outperforms the competitors, see \cref{sec:random_results_appendix}.
}

\change{
Finally, we analyze the disentanglement of the learned concepts. 
Thanks to expert annotation, \method with both backbones fares better than all competitors across datasets, with a sensible improvement in \SHAPES and \DERMA. On \CUB, \method ranks first with \DINO ($\approx 0.71$) but shows lower performance with \CLIP ($\approx 0.53$), while \VLGCBM is the second-best method (with \Disent$\approx0.61$).
The propagation of expert annotation through the GP backbone of \method shows widespread improvements in \Disent compared to \CBM{@}.%
}
Overall, these results indicate that propagating expert supervision with our method can improve concept quality not only in accuracy but also in terms of calibration and disentanglement \change{(even though for the latter the benefit is backbone-dependent across datasets)}.

\textbf{Q3: \method achieves competitive task performance}. 
\cref{tab:results} shows that our method outperforms the VLM-CBMs (\LABO, \LFCBM, and \VLGCBM) in predicting the task labels on all datasets, \change{exception made for \LFCBM in \CelebA}. 
The gap between \method and VLM-based methods is particularly evident on \CUB \change{($+21$ p.p. wrt the runner up) and \DERMA ($+12$ p.p.)}, while \CelebA is the only dataset where all methods achieve similar results in terms of \FY.
\change{When comparing \method with \texttt{CBM}-@ and \CBM{-@100}, \method achieves higher or similar \FY, winning in \CelebA and \CUB, while performing on par in \SHAPES and \DERMA.
In \CUB, we notice that the choice of the backbone affects sensibly the \FY for \method (around $8$ p.p. of gain when using \texttt{DINO} instead of \CLIP).
}

This shows that combining VLM representations with a limited amount of human supervision greatly preserve the relevant information for the final task, even surpassing \CBM{@100} in \CUB and \CelebA.

\section{Related Work}
\label{sec:related-work}

\textbf{Self-explainable Neural Networks.}  Explainable-by-design neural networks have garnered considerable popularity \citep{poeta2023concept, ji2025comprehensive}, owing to their promise of high task accuracy and interpretability.  In all these approaches, the flow of information from input to prediction is mediated by high-level concepts learned from data.
\change{Learning with only task labels supervision does not ensure the model's concepts are aligned with human semantics \citep{bortolotti2025shortcuts}, 
and known conditions to provably recover the concepts (\eg $1$-sparsity of the concepts \citep{goyal2025causal}, access to multiple environments \citep{zheng2025nonparametric}) might not hold at scale.
CBMs \citep{koh2020concept} \change{and their upgrades \citep{zarlenga2022concept, sawada2022concept, havasi2022addressing, marconato2022glancenets, kim2023probabilistic, vandenhirtz2024stochastic}} supervise the concept bottleneck during training, cf. \cref{sec:preliminaries}, \change{and as such suffer from impaired applicability (\textbf{D3}).}}
Recent VLM-CBMs improve on this aspect by replacing expert annotations with a VLM's \citep{oikarinen2023label, yang2023language, yuksekgonul2023post, lai2023faithful,
kazmierczak2024clipqdaexplainableconceptbottleneck, 
rao2024discover, knab2024dcbm}, but in doing so risk lowering concept accuracy (\textbf{D2}), as shown in \citep{debole2025if} and in \cref{sec:experiments}.
\method{s} combine the best of both worlds:  global domain-specific information from the VLMs embedding space and local information from expert annotations.  
\change{We exclude VLM-CBMs that focus primarily on concept discovery (\eg \citep{knab2024dcbm, rao2024discover}), as they automatically identify the bottleneck concepts rather than relying on predefined ones, and therefore lack ground-truth concept annotations, making it difficult to measure concept accuracy and calibration.}

\change{\textbf{Gaussian Processes in embedding space}.}
The works most closely related to ours are two.  \citet{feng2024bayesian} employ GPs to decide when to add new concepts in the vocabulary.  This is orthogonal to our work and could be combined with it.
\citet{wang2023gaussian}, like us, build on the GP architecture of \citep{milios2018dirichlet}, to obtain confidence estimates for concept activations to be used for explaining the model.  This technique is compatible with \method, although our method is mainly designed for propagating expert supervision instead \change{and leverages expert annotations in an active manner}.

\textbf{Concept quality and interpretability}.  While inaccurate concepts cannot match human semantics, achieving high concept accuracy might not guarantee full interpretability of the bottleneck \citep{margeloiu2021concept, mahinpei2021promises, marconato2023interpretability}. 
Recent works have shown that CBMs can maximize task performance by learning concepts that depend on spurious information, such as unrelated concepts -- causing \textit{entanglement} \citep{higgins2018towards, kazhdan2021disentanglement, scholkopf2021toward, suter2019robustly, eastwood2018framework} -- and contextual or stylistic cues -- causing \textit{information leakage} \citep{margeloiu2021concept, mahinpei2021promises, marconato2022glancenets, havasi2022addressing} and sensitivity to \textit{non-local changes} of the concept bottleneck \citep{raman2023concept, srivastava2024vlgcbm}.
While resorting to VLM-CBMs favors better applicability and promotes the access to rich embedding spaces, it hides interpretability issues leading to high leakage and low disentanglement in practice \citep{srivastava2024vlgcbm, debole2025if}.

\section{Conclusion}
\label{sec:conclusion}

We introduce \method, a simple yet effective concept-based approach that strikes a middle ground between interpretability and applicability. \method infers concept activations by \cchange{(optionally, actively) collecting and} propagating expert concept supervision according to the geometry of a VLM’s embedding space. Our results show that, with only a small amount of concept supervision, \method achieves substantial improvements in concept accuracy \change{and calibration} over standard VLM-CBMs. \change{Well-calibrated concept predictions naturally enable active learning, allowing efficient selection of the most informative samples to annotate and further reducing the supervision burden. Finally, the improved concept accuracy yields more semantically aligned concepts and better downstream task performance.}

While \method{s} are designed to rapidly propagate concept annotations, thus lessening the cost of traditional CBMs, pairing it with losses that promote alignment and discourage entanglement is an open future venue.  We expect that \method{s} may also benefit by other post-hoc correction routines on latent embeddings that build on disentanglement analyses, \eg \citep{kazmierczak2025enhancing}, and structured latent spaces based on object-centric representations, \eg  \citep{steinmann2025object}.
In the same vein, another promising line of research involves exploiting the VLM's embedding space, which spans outside the application domain, to construct out-of-distribution (OOD) acquisition strategies tailored \change{for improving concept quality by} removing entanglement and \change{concept} leakage \citep{margeloiu2021concept, mahinpei2021promises}, \change{as doing so requires OOD data in general \citep{marconato2023interpretability}.}

\newpage

\bibliography{paper}

\appendix

\section{Pipeline}
\label{sec:pipeline-appendix}

\begin{figure*}[h]
    \centering
    \includegraphics[width=\linewidth]{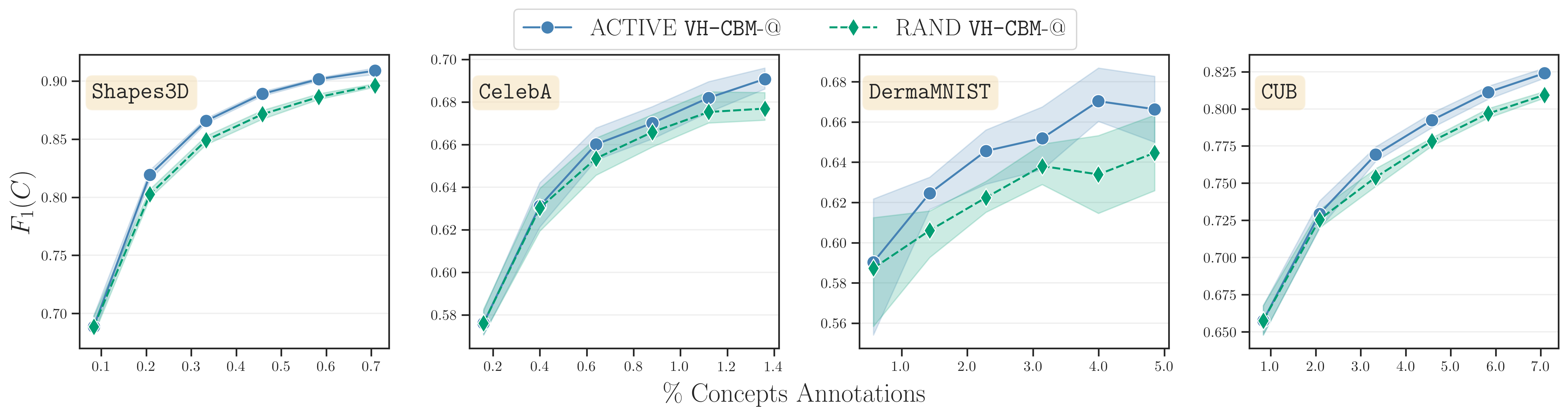}

    \includegraphics[width=\linewidth]{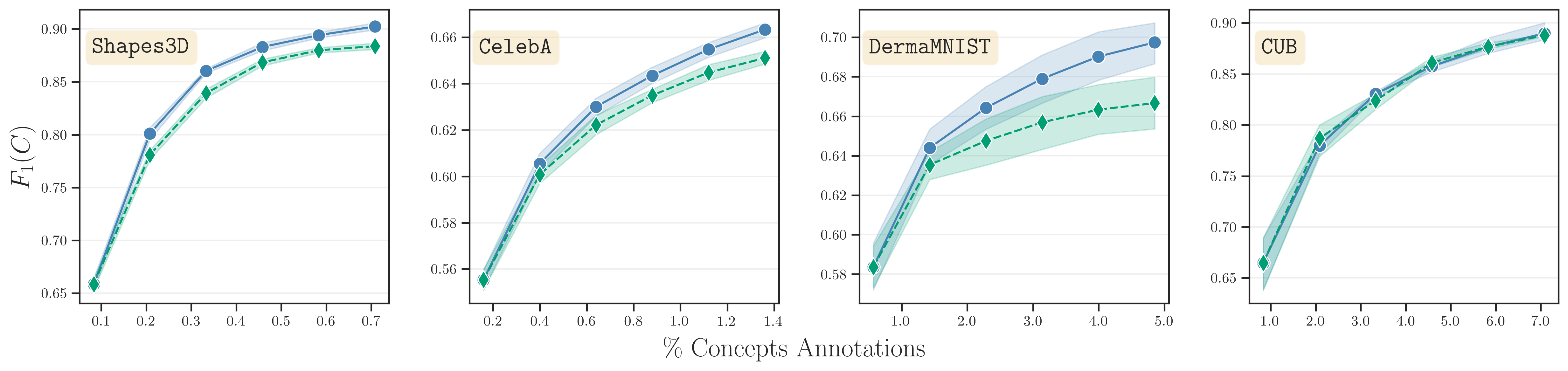}
    \caption{Comparison between \texttt{Active} vs \texttt{Random} acquisition rule for the selection of new concepts to annotate. We tested two different backbones: \CLIP (top) and \DINO (bottom). While \texttt{Random} takes a pre-determined number of random concepts to annotate, \texttt{Active} uses the GP uncertainty to select which concepts to annotate.}
    \label{fig:active-vs-rand-clip}
    \label{fig:active-vs-rand-dino}
\end{figure*}

\change{The choice of the number of annotated samples to start with, as well as the number of concepts to annotate at each step, is arbitrary. We choose 40 as the initial value to keep annotation costs low while providing enough samples for the GP to begin propagating information. We tested other values with consistent results. The same applies to the step size: the number of concepts annotated per iteration can be adjusted to reflect different user preferences. If minimal annotation effort is desired, smaller step sizes are preferable; if faster convergence is the priority, larger values can be used to reach good performance in fewer iterations.
At each iteration, the desired number of samples is collected and annotated. The selection of which concepts to annotate is based on an active acquisition function: we query the GP's uncertainty estimates and prioritize the most uncertain concepts for annotation, leaving the remaining ones unannotated. Alternatively, concepts can be selected at random, which also yields competitive results. We compare the active and random acquisition functions in \cref{fig:active-vs-rand-clip}.}

\subsection{Uncertainty-based Acquisition Function}
\label{sec:active-selection}

\change{More specifically,} our active acquisition function begins by randomly sampling a subset of the training data. This promotes exploration and avoids clustering annotations around uncertain samples that are nonetheless close to one another in the embedding space, which would be inefficient in terms of annotation cost. Within this subset, we measure the GP uncertainty and annotate only the concepts with high uncertainty, effectively discarding the least uncertain candidates.

\cchange{
For instance, consider \texttt{CelebA}, where each image has 39 concepts. In the random baseline, 60 images are drawn and all their concepts annotated, yielding 2340 concept annotations per iteration. In our active selection procedure, we instead draw a larger pool of 95 images (3705 concepts in total), measure the GP uncertainty for each concept, and select the 2340 most uncertain concept-image pairs. This way, both strategies share the same annotation budget, but active selection avoids redundant annotations on concepts the model is already confident about, allocating the budget where it is most informative.

The size of the random pool can be adjusted, governing the exploration-exploitation trade-off: a larger pool allows more selective concept annotation, but also increases the risk of drawing uninformative samples. This choice is ultimately data-dependent; in all our experiments we set it to 95 images, as in the \texttt{CelebA} example above, and we observed that small variations around this value do not significantly affect performance.}

\section{Random concept selection \method}
\label{sec:random_results_appendix}

\begin{figure*}[h]
    \centering
    \includegraphics[width=\linewidth]{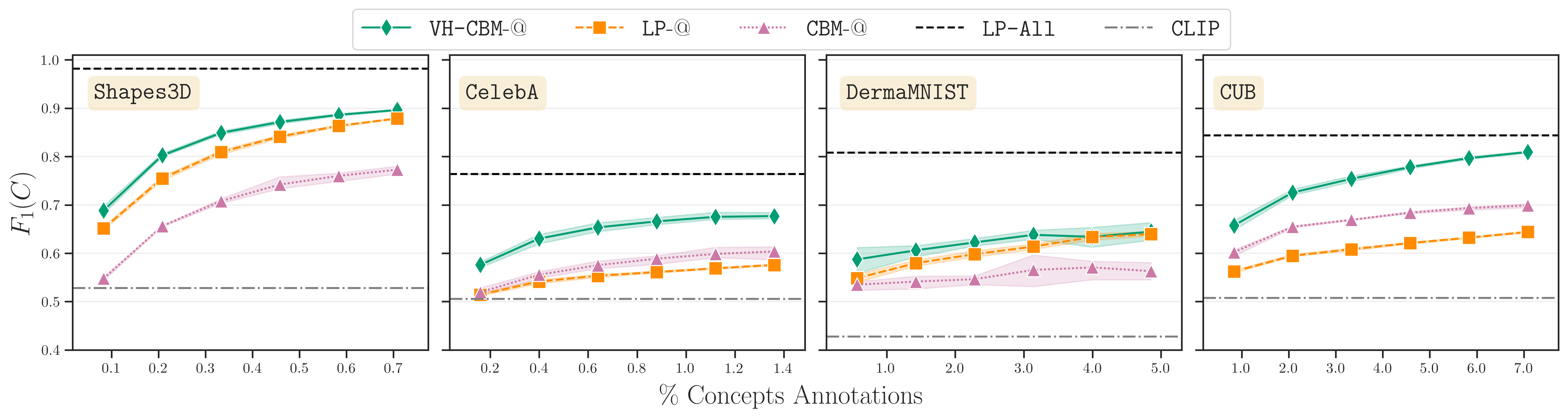}

    \includegraphics[width=\linewidth]{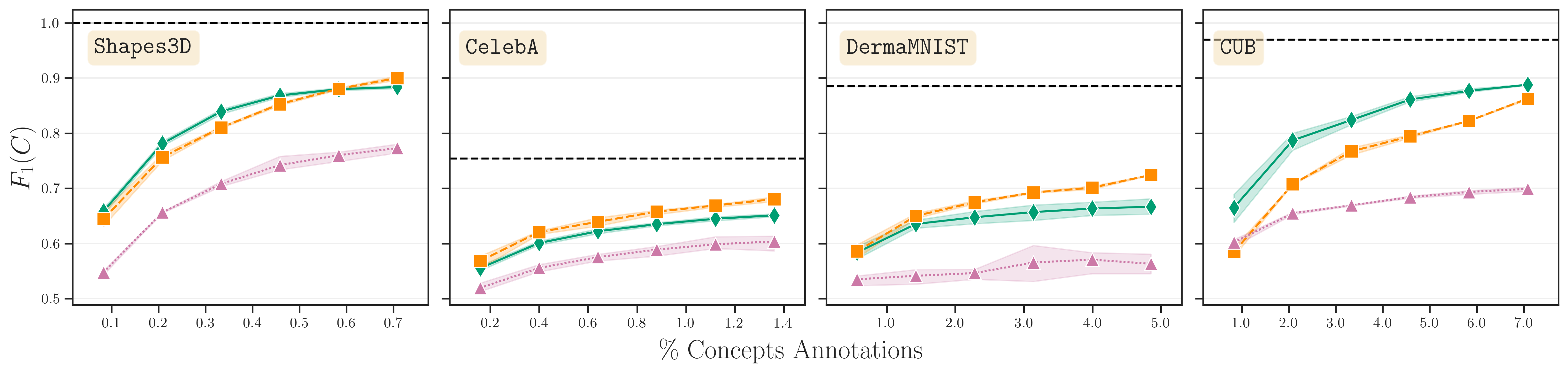}
    \caption{{\change{\method improves concept accuracy also with \texttt{Random} annotation strategy.}}
    \FC of \method with varying percentages of concept supervision for the \CLIP (top) and \DINO (bottom) backbones.  Results are averaged over three runs, with confidence intervals estimated via bootstrap resampling across runs.
    Random acquisition with \method surpasses linear probing at early regimes in $6$ to $8$ cases, and wins at the last step in $4$ cases, while being competitive to linear probing in the remaining ones. 
    }
    \label{fig:fig2-ablation-random-clip}
    \label{fig:fig2-ablation-random-dino}
\end{figure*}

We report results for the random annotation baseline, where at each iteration 60 samples are fully annotated. As shown in \cref{tab:random-results-appendix} and \cref{fig:fig2-ablation-random-clip}, random selection already yields consistent improvements, confirming that even without active sampling, \method benefits reliably from additional supervision.

\begin{table*}[!t]
    \centering
    \caption{
    Comparison when \method adopts the \texttt{Random} annotation strategy. Here,
    {\method{@} remains more accurate \change{and calibrated}} \change{compared to other VLM-CBMs and \CBM{@} across datasets when trained on $340$ randomly concept annotated samples}.  Best is in \textbf{bold}, second-best \underline{underlined}. 
    }
    \scalebox{0.85}{
    \begin{tabular}{llccccc}
        \toprule
        & {\sc Model} & \FY ($\uparrow$) & \FC ($\uparrow$) & \ROCAUC ($\uparrow$) & \Disent ($\uparrow$) & \ECER ($\downarrow$) \\
        \cmidrule{1-7}
        \multirow[c]{7}{*}{\rotatebox{90}{\SHAPES}}
        & \CBM @ $100\%$
            & $0.99 \pm 0.01$     %
            & $0.99 \pm 0.01$     %
            & $0.99 \pm 0.01$ %
            & $0.99 \pm 0.01$   
            & $0.38 \pm 0.07$   \\ %
                    \cmidrule{2-7}
        & \cellcolor{gray!10}\CBM @  $0.71\%$
            & \cellcolor{gray!10}$0.94 \pm 0.01$     %
            & \cellcolor{gray!10}$0.77 \pm 0.01$     %
            & \cellcolor{gray!10}$0.92 \pm 0.01$ %
            & \cellcolor{gray!10}$\underline{0.63 \pm 0.01}$   
            & \cellcolor{gray!10}$0.39 \pm 0.07$   \\ %

        & \LABO
            & $0.76 \pm 0.01$     %
            & $0.56 \pm 0.01$     %
            & $0.72 \pm 0.01$ %
            & $0.28 \pm 0.01$   
            & $0.89 \pm 0.02$   \\ %
        & \cellcolor{gray!10}\LFCBM
            & \cellcolor{gray!10}$0.96 \pm 0.01$     %
            & \cellcolor{gray!10}$0.57 \pm 0.01$     %
            & \cellcolor{gray!10}$0.72 \pm 0.01$ %
            & \cellcolor{gray!10}$0.28 \pm 0.01$   
            & \cellcolor{gray!10}$0.39 \pm 0.29$   \\ %
        & \VLGCBM
            & $0.49 \pm 0.06$     %
            & $0.54 \pm 0.01$     %
            & $0.68 \pm 0.01$ %
            & $0.24 \pm 0.01$   
            & $\underline{0.31 \pm 0.25}$   \\ %

         & \cellcolor{blue!10}\method \texttt{CLIP} @ $0.71\% $
            & \cellcolor{blue!10}$\mathbf{0.99 {\pm 0.01}}$     %
            & \cellcolor{blue!10}$\underline{0.89 \pm 0.01}$     %
            
            & \cellcolor{blue!10}$\underline{0.95 \pm 0.01}$ %
            & \cellcolor{blue!10}$\mathbf{0.76 {\pm 0.03}}$   %
            & \cellcolor{blue!10}$\mathbf{0.03 \pm 0.01}$   \\ %
            
        & \cellcolor{blue!10}\method \texttt{DINO} @ $0.71\% $
            & \cellcolor{blue!10}$\underline{0.98 \pm 0.01}$     %
            & \cellcolor{blue!10}$\mathbf{0.91 {\pm 0.01}}$     %
            
            & \cellcolor{blue!10}$\mathbf{0.97 {\pm 0.01}}$ %
            & \cellcolor{blue!10}$\mathbf{0.76 {\pm 0.03}}$    %
            & \cellcolor{blue!10}$\mathbf{0.03 \pm 0.01}$   \\ %
            
        \midrule\midrule

        \multirow[c]{7}{*}{\rotatebox{90}{\CelebA}}
        & \CBM @ $100\%$
            & $0.96 \pm 0.01$     %
            & $0.77 \pm 0.01$     %
            & $0.93 \pm 0.01$ %
            & $0.75 \pm 0.01$   
            & $0.32 \pm 0.15$   \\ %
                    \cmidrule{2-7}
        & \cellcolor{gray!10}\CBM @ $1.4\%$
            & \cellcolor{gray!10} $0.90 \pm 0.01$     %
            & \cellcolor{gray!10}$0.60 \pm 0.01$     %
            & \cellcolor{gray!10}$\underline{0.81 \pm 0.01}$ %
            & \cellcolor{gray!10}$\underline{0.37 \pm 0.02}$   
            & \cellcolor{gray!10}$\underline{0.31 \pm 0.14}$   \\ %
        & \LABO
            & $\underline{0.98 \pm 0.01}$     %
            & $0.55 \pm 0.01$     %
            & $0.66 \pm 0.01$ %
            & $0.28 \pm 0.01$   
            & $0.77 \pm 0.20$   \\ %
        & \cellcolor{gray!10}\LFCBM
            & \cellcolor{gray!10}$\mathbf{0.99 \pm 0.01}$     %
            & \cellcolor{gray!10}$0.58 \pm 0.01$     %
            & \cellcolor{gray!10}$0.68 \pm 0.01$ %
            & \cellcolor{gray!10}$0.29 \pm 0.01$   
            & \cellcolor{gray!10}$0.40 \pm 0.24$   \\ %
        & \VLGCBM
            & $\underline{0.98 \pm 0.01}$     %
            & $0.54 \pm 0.01$     %
            & $0.62 \pm 0.01$ %
            & $0.22 \pm 0.01$   
            & $0.42 \pm 0.27$   \\ %
            
        & \cellcolor{blue!10}\method \texttt{CLIP} @ $1.4\%$
            &\cellcolor{blue!10}$\underline{0.98 \pm 0.01}$     %
            &\cellcolor{blue!10} $\mathbf{0.68 {\pm 0.01}}$     %
            & \cellcolor{blue!10}$\mathbf{0.83 {\pm 0.01}}$ %
            & \cellcolor{blue!10}$0.35 \pm 0.08$    %
            & \cellcolor{blue!10}$\mathbf{0.05 \pm 0.03}$   \\ %
            
        & \cellcolor{blue!10}\method \texttt{DINO} @ $1.4\%$
            & \cellcolor{blue!10}$0.97 \pm 0.01$     %
            & \cellcolor{blue!10}$\underline{0.67 \pm 0.01}$     %
            
            & \cellcolor{blue!10}$\mathbf{0.83 \pm 0.01}$ %
            & \cellcolor{blue!10}$\mathbf{0.42 \pm 0.01}$    %
            & \cellcolor{blue!10}$\mathbf{0.05 \pm 0.03}$   \\ %

        \midrule\midrule %

        \multirow[c]{7}{*}{\rotatebox{90}{\DERMA}}
        & \CBM @ $100\%$
            & $0.68 \pm 0.03$     %
            & $0.65 \pm 0.01$     %
            & $0.91 \pm 0.00$ %
            & $0.28 \pm 0.03$   
            & $0.41 \pm 0.11$   \\ %
                    \cmidrule{2-7}
        & \cellcolor{gray!10}\CBM @ $4.8\%$
           & \cellcolor{gray!10}$\underline{0.65 \pm 0.05}$     %
            & \cellcolor{gray!10}$0.56 \pm 0.02$     %
            & \cellcolor{gray!10}$0.80 \pm 0.02$ %
            & \cellcolor{gray!10}$0.15 \pm 0.03$   
            & \cellcolor{gray!10}$0.40 \pm 0.13$   \\ %
        & \LABO
            & ${0.57 \pm 0.01}$     %
            & $0.42 \pm 0.01$     %
            & $0.57 \pm 0.01$ %
            & $0.04 \pm 0.01$   
            & $0.86 \pm 0.22$   \\ %
        & \cellcolor{gray!10}\LFCBM
            & \cellcolor{gray!10}$0.51 \pm 0.03$     %
            & \cellcolor{gray!10}$0.42 \pm 0.01$     %
            & \cellcolor{gray!10}$0.56 \pm 0.01$ %
            & \cellcolor{gray!10}$0.05 \pm 0.01$   
            & \cellcolor{gray!10}$\underline{0.14 \pm 0.22}$   \\ %
        & \VLGCBM
            & $0.43 \pm 0.01$     %
            & $0.42 \pm 0.01$     %
            & $0.50 \pm 0.01$ %
            & $0.02 \pm 0.01$   
            & $0.27 \pm 0.35$   \\ %

        & \cellcolor{blue!10}\method \texttt{CLIP} @ $4.8\%$
            & \cellcolor{blue!10}$0.63 \pm 0.01$     %
            & \cellcolor{blue!10}$\underline{0.64 \pm 0.01}$     %
            & \cellcolor{blue!10}$\underline{0.86 \pm 0.01}$ %
            & \cellcolor{blue!10}$\underline{0.23 \pm 0.06}$    %
            & \cellcolor{blue!10}$\mathbf{0.02 \pm 0.01}$   \\ %

        & \cellcolor{blue!10}\method \texttt{DINO} @ $4.8\%$
            & \cellcolor{blue!10}$\mathbf{0.66 \pm 0.01}$     %
            & \cellcolor{blue!10}$\mathbf{0.70 \pm 0.01}$     %
            & \cellcolor{blue!10}$\mathbf{0.91 \pm 0.01}$ %
            & \cellcolor{blue!10}$\mathbf{0.28 \pm 0.01}$    %
            & \cellcolor{blue!10}$\mathbf{0.02 \pm 0.01}$   \\ %

        \midrule\midrule

        \multirow[c]{7}{*}{\rotatebox{90}{\CUB}}
        & \CBM @ $100\%$
            & $0.69 \pm 0.01$     %
            & $0.86 \pm 0.01$     %
            & $0.97 \pm 0.01$ %
            & $0.75 \pm 0.01$   
            & $0.32 \pm 0.13$   \\ %
                    \cmidrule{2-7}
        & \cellcolor{gray!10}\CBM @ $7.0\%$
            & \cellcolor{gray!10} $0.63 \pm 0.01$     %
            & \cellcolor{gray!10}$0.70 \pm 0.01$     %
            & \cellcolor{gray!10}$0.84 \pm 0.01$ %
            & \cellcolor{gray!10}$0.37 \pm 0.01$   
            & \cellcolor{gray!10}$0.32 \pm 0.12$   \\ %
        & \LABO
            & $0.27 \pm 0.01$     %
            & $0.56 \pm 0.01$     %
            & $0.66 \pm 0.01$ %
            & $0.19 \pm 0.01$   
            & $0.80 \pm 0.13$   \\ %
        & \cellcolor{gray!10}\LFCBM
            & \cellcolor{gray!10}$0.67 \pm 0.01$     %
            & \cellcolor{gray!10}$0.48 \pm 0.01$     %
            & \cellcolor{gray!10}$0.45 \pm 0.01$ %
            & \cellcolor{gray!10}$0.19 \pm 0.01$   
            & \cellcolor{gray!10}$\underline{0.21 \pm 0.09}$   \\ %
        & \VLGCBM
            & $0.60 \pm 0.01$     %
            & $0.79 \pm 0.01$     %
            & $0.91 \pm 0.01$ %
            & $\underline{0.61 \pm 0.01}$   
            & $0.23 \pm 0.13$   \\ %

        & \cellcolor{blue!10}\method \texttt{CLIP} @ $7.0\%$
            & \cellcolor{blue!10}$\underline{0.79 \pm 0.01}$     %
            & \cellcolor{blue!10}$\underline{0.81 \pm 0.01}$     %
            & \cellcolor{blue!10}$\underline{0.92 \pm 0.04}$ %
            & \cellcolor{blue!10}$0.54 \pm 0.02$    %
            & \cellcolor{blue!10}$\mathbf{0.02 \pm 0.01}$   \\ %
            
        & \cellcolor{blue!10}\method \texttt{DINO} @ $7.0\%$
            & \cellcolor{blue!10}$\mathbf{0.89 \pm 0.01}$     %
            & \cellcolor{blue!10}$\mathbf{0.89 \pm 0.01}$     %
            & \cellcolor{blue!10}$\mathbf{0.96 \pm 0.01}$ %
            & \cellcolor{blue!10}$\mathbf{0.68 \pm 0.01}$    %
            & \cellcolor{blue!10}$\mathbf{0.02 \pm 0.01}$   \\ %
            
        \bottomrule
    \end{tabular}
    }
    \label{tab:random-results-appendix}
\end{table*}

\section{Calibration}
\label{sec:calibration-appendix}
To measure calibration, we evaluate both the baseline and competitor models using the ECCE metric \citep{arrieta2022metrics}. Although the Expected Calibration Error (ECE) is the standard measure of calibration, it is heavily dependent on the choice of binning; the ECCE metric is specifically designed to overcome this limitation. 
In the results reported in \cref{tab:results} we adopt the ECCE-R variant, which measures the full range of the cumulative deviations.
As in \citep{arrieta2022metrics}, cumulative deviations $C_k$ are defined as
\begin{equation}
    C_k = \frac{1}{n} \sum_{j=1}^k (R_j - S_j)
\end{equation}
where $S_j$ are the predicted (sorted) scores and $R_j$ the actual outcomes. ECCE-MAD is the maximum absolute value of the cumulative differences across all observations; ECE1 corresponds to the weighted sum of the absolute values of the differences between the average response and the average score in each bin, while ECE2 is the weighted sum of the squares of the differences between the average response and the average score in each bin.

For completeness, a table reporting all calibration metrics is also provided in \cref{tab:full-calibration-results}.
It should be noted that calibration is measured with respect to the concept predictions — that is, the output of the backbone — within each CBM variant, rather than with respect to the final label predictions.
\change{We observe that \method gives the best calibration results across datasets, except in \SHAPES where it ranks second-best and \texttt{LP-@} demonstrates particularly competitive. \texttt{LP-@} fares good also on \DERMA (it is second best on all columns), but sensibly downgrades w.r.t. \method on \CelebA and \CUB. Other supervised methods do not compete with \method, which shows low calibration scores and improved concept accuracy compared to VLM-CBMs. On calibration \LABO, \LFCBM, and \VLGCBM lag behind \method with seldom good results; \eg \LFCBM in \DERMA fares good, but lacks in all other datasets.
}

\begin{table*}[h]
    \centering
    \caption{\textbf{Extensive calibration evaluation} over competitors and baselines, including calibration on linear probes (\texttt{LP}-@) on \CLIP embeddings, using the metrics from \citep{arrieta2022metrics}. Best is in \textbf{bold}, second-best \underline{underlined}.}
    \scalebox{0.85}{
    \begin{tabular}{llcccc}
        \toprule
        & {\sc Method} & \textsc{ECCE-R} ($\downarrow$) & \textsc{ECCE-MAD} ($\downarrow$) & \textsc{ECE1} ($\downarrow$) & \textsc{ECE2} ($\downarrow$) \\
        \cmidrule{1-6}

        \multirow[c]{8}{*}{\rotatebox{90}{\SHAPES}} 
         & \CBM @ $100\%$ & $0.38 \pm 0.07$ & $0.38 \pm 0.07$ & $0.38 \pm 0.07$ & $0.15 \pm 0.05$ \\
         & \cellcolor{gray!10}\texttt{CBM}-@ $0.71\%$ &\cellcolor{gray!10} $0.39 \pm 0.07$ &\cellcolor{gray!10} $0.39 \pm 0.07$ &\cellcolor{gray!10} $0.39 \pm 0.07$ &\cellcolor{gray!10} $0.16 \pm 0.05$ \\
         & \texttt{LP}-@ $0.71\%$& $\mathbf{0.02 \pm 0.03}$ & $\mathbf{0.01 \pm 0.02}$ & $\mathbf{0.03 \pm 0.04}$ & $\mathbf{0.01 \pm 0.01}$ \\
         &\cellcolor{gray!10}\texttt{\LABO} &\cellcolor{gray!10} $0.89 \pm 0.02$ &\cellcolor{gray!10} $0.89 \pm 0.02$ &\cellcolor{gray!10} $0.89 \pm 0.02$ &\cellcolor{gray!10} $0.79 \pm 0.04$ \\
         & \texttt{\LFCBM} & $0.39 \pm 0.29$ & $0.37 \pm 0.30$ & $0.42 \pm 0.27$ & $0.30 \pm 0.26$ \\

         &\cellcolor{gray!10}\texttt{VLGCBM} &\cellcolor{gray!10} $0.31 \pm 0.25$ &\cellcolor{gray!10} $0.29 \pm 0.26$ &\cellcolor{gray!10} $0.33 \pm 0.24$ &\cellcolor{gray!10} $0.21 \pm 0.22$ \\
         & \cellcolor{blue!10}\method \texttt{CLIP} @ $0.71\%$& \cellcolor{blue!10}$\underline{0.03 \pm 0.01}$ & \cellcolor{blue!10}$\underline{0.02 \pm 0.01}$ & \cellcolor{blue!10}$\underline{0.05 \pm 0.02}$ & \cellcolor{blue!10}$\mathbf{0.01 \pm 0.01}$ \\
         & \cellcolor{blue!10}\method \DINO @ $0.71\%$ & \cellcolor{blue!10}$0.04 \pm 0.01$ & \cellcolor{blue!10}$0.03 \pm 0.01$ & \cellcolor{blue!10}$0.07 \pm 0.03$ & \cellcolor{blue!10}$\underline{0.02 \pm 0.01}$ \\
        \midrule\midrule
        
        \multirow[c]{8}{*}{\rotatebox{90}{\CelebA}} 
         & \CBM @ $100\%$ & $0.32 \pm 0.15$ & $0.32 \pm 0.15$ & $0.32 \pm 0.15$ & $0.13 \pm 0.09$ \\
         &\cellcolor{gray!10}\texttt{CBM}-@ $1.4\%$&\cellcolor{gray!10} $0.31 \pm 0.14$ &\cellcolor{gray!10} $0.31 \pm 0.14$ &\cellcolor{gray!10} $0.31 \pm 0.14$ &\cellcolor{gray!10} $0.12 \pm 0.08$ \\
         & \texttt{LP}-@ $1.4\%$& $0.28 \pm 0.09$ & $0.25 \pm 0.10$ & $0.32 \pm 0.08$ & $0.20 \pm 0.10$ \\
         &\cellcolor{gray!10}\texttt{\LABO} &\cellcolor{gray!10} $0.77 \pm 0.20$ &\cellcolor{gray!10} $0.77 \pm 0.20$ &\cellcolor{gray!10} $0.77 \pm 0.20$ &\cellcolor{gray!10} $0.64 \pm 0.27$ \\
         & \texttt{\LFCBM} & $0.40 \pm 0.24$ & $0.39 \pm 0.24$ & $0.44 \pm 0.22$ & $0.30 \pm 0.24$ \\

         &\cellcolor{gray!10}\texttt{\VLGCBM} &\cellcolor{gray!10} $0.42 \pm 0.27$ &\cellcolor{gray!10} $0.42 \pm 0.27$ &\cellcolor{gray!10} $0.43 \pm 0.27$ &\cellcolor{gray!10} $0.27 \pm 0.28$ \\
         & \cellcolor{blue!10}\method \texttt{CLIP} @ $1.4\%$& \cellcolor{blue!10}$\mathbf{0.04 \pm 0.03}$ & \cellcolor{blue!10}$\mathbf{0.04 \pm 0.03}$ & \cellcolor{blue!10}$\underline{0.06 \pm 0.04}$ & \cellcolor{blue!10}$\mathbf{0.01 \pm 0.01}$ \\
         & \cellcolor{blue!10}\method \DINO @ $1.4\%$& \cellcolor{blue!10}$\mathbf{0.04 \pm 0.03}$ & \cellcolor{blue!10}$\mathbf{0.04 \pm 0.03}$ & \cellcolor{blue!10}$\mathbf{0.05 \pm 0.04}$ & \cellcolor{blue!10}$\mathbf{0.01 \pm 0.01}$ \\
        \midrule\midrule

        \multirow[c]{8}{*}{\rotatebox{90}{\DERMA}} 
         & \CBM @ $100\%$ & $0.41 \pm 0.11$ & $0.40 \pm 0.11$ & $0.41 \pm 0.11$ & $0.18 \pm 0.07$ \\
         &\cellcolor{gray!10}\texttt{CBM}-@ $4.8\%$&\cellcolor{gray!10} $0.40 \pm 0.13$ &\cellcolor{gray!10} $0.40 \pm 0.13$ &\cellcolor{gray!10} $0.40 \pm 0.13$ &\cellcolor{gray!10} $0.18 \pm 0.08$ \\
         & \texttt{LP}-@ $4.8\%$& $0.07 \pm 0.06$ & $0.06 \pm 0.06$ & $0.08 \pm 0.07$ & $0.03 \pm 0.03$ \\
         &\cellcolor{gray!10}\texttt{\LABO} &\cellcolor{gray!10} $0.86 \pm 0.22$ &\cellcolor{gray!10} $0.86 \pm 0.22$ &\cellcolor{gray!10} $0.86 \pm 0.22$ &\cellcolor{gray!10} $0.78 \pm 0.29$ \\
         & \texttt{\LFCBM} & $0.14 \pm 0.22$ & $0.14 \pm 0.22$ & $0.14 \pm 0.22$ & $0.07 \pm 0.16$ \\

         &\cellcolor{gray!10}\texttt{\VLGCBM} &\cellcolor{gray!10} $0.27 \pm 0.35$ &\cellcolor{gray!10} $0.27 \pm 0.35$ &\cellcolor{gray!10} $0.27 \pm 0.35$ &\cellcolor{gray!10} $0.19 \pm 0.34$ \\
         & \cellcolor{blue!10}\method \texttt{CLIP} @ $4.8\%$ & \cellcolor{blue!10}$\mathbf{0.02 \pm 0.01}$ & \cellcolor{blue!10}$\mathbf{0.02 \pm 0.01}$ & \cellcolor{blue!10}$\mathbf{0.02 \pm 0.01}$ & \cellcolor{blue!10}$\mathbf{0.01 \pm 0.01}$ \\
         & \cellcolor{blue!10}\method \DINO @ $4.8\%$& \cellcolor{blue!10}$\mathbf{0.02 \pm 0.01}$ & \cellcolor{blue!10}$\mathbf{0.02 \pm 0.01}$ & \cellcolor{blue!10}$\mathbf{0.03 \pm 0.01}$ & \cellcolor{blue!10}$\mathbf{0.01 \pm 0.01}$ \\
        \midrule\midrule
        
        \multirow[c]{8}{*}{\rotatebox{90}{\CUB}} 
         & \CBM @ $100\%$ & $0.32 \pm 0.13$ & $0.32 \pm 0.13$ & $0.32 \pm 0.13$ & $0.12 \pm 0.08$ \\
         &\cellcolor{gray!10}\texttt{CBM}-@ $7.0\%$&\cellcolor{gray!10} $0.32 \pm 0.12$ &\cellcolor{gray!10} $0.32 \pm 0.13$ &\cellcolor{gray!10} $0.32 \pm 0.12$ &\cellcolor{gray!10} $0.12 \pm 0.07$ \\
         & \texttt{LP}-@ $7.0\%$& $0.23 \pm 0.08$ & $0.21 \pm 0.08$ & $0.26 \pm 0.06$ & $0.15 \pm 0.08$ \\
         &\cellcolor{gray!10}\texttt{\LABO} &\cellcolor{gray!10} $0.80 \pm 0.13$ &\cellcolor{gray!10} $0.80 \pm 0.13$ &\cellcolor{gray!10} $0.80 \pm 0.13$ &\cellcolor{gray!10} $0.66 \pm 0.19$ \\
         & \texttt{\LFCBM} & $0.21 \pm 0.09$ & $0.18 \pm 0.10$ & $0.26 \pm 0.10$ & $0.13 \pm 0.08$ \\

         &\cellcolor{gray!10}\texttt{\VLGCBM} &\cellcolor{gray!10} $0.23 \pm 0.13$ &\cellcolor{gray!10} $0.23 \pm 0.13$ &\cellcolor{gray!10} $0.24 \pm 0.13$ &\cellcolor{gray!10} $0.12 \pm 0.11$ \\
         & \cellcolor{blue!10}\method \texttt{CLIP} @ $7.0\%$& \cellcolor{blue!10}$\mathbf{0.02 \pm 0.01}$ & \cellcolor{blue!10}$\mathbf{0.02 \pm 0.01}$ & \cellcolor{blue!10}$\underline{0.03 \pm 0.01}$ & \cellcolor{blue!10}$\mathbf{0.01 \pm 0.01}$ \\
         & \cellcolor{blue!10}\method \DINO @ $7.0\%$ & \cellcolor{blue!10}$\mathbf{0.02 \pm 0.01}$ & \cellcolor{blue!10}$\mathbf{0.02 \pm 0.01}$ & \cellcolor{blue!10}$\mathbf{0.02 \pm 0.01}$ & \cellcolor{blue!10}$\mathbf{0.01 \pm 0.01}$ \\
        \bottomrule
    \end{tabular}
    }
    \label{tab:full-calibration-results}
\end{table*}

\section{Implementation Details}
\label{sec:appendix-impl-details}

We implement the GP using \texttt{GPytorch} \citep{gardner2018gpytorch}.
In order to scale to larger annotated sets, we employ a sparse GP that leverages a smaller set of inducing points to approximate a full GP distribution over the data \citep{snelson2005sparse}, implemented through \verb|gpytorch.models.ApproximateGP|.
As mentioned in \cref{sec:method}, we use a constant mean function (\verb|gpytorch.means.ConstantMean|) and an RBF kernel (\verb|gpytorch.kernels.RBFKernel|).

\subsection{Hyperparameters}
\label{sec:hyperparameters}

In terms of the GP hyperparameters, for every latent GP we initialize \textit{length scale} $\rho_{ij}$ to $1.41$, the \textit{output scale} $\alpha_{ij}$ to $1.0$ and the mean function to $0.0$. These are learned during training in order to find the optimal values.
The noise for the Dirichlet Likelihood is set as $0.01$ and
the number of inducing points is set to that of annotated samples (\ie images that have expert annotations available).

\subsection{Running times}
\label{sec:compute-resources}
The training time for reaching convergence in a single iteration step of the multi-label multi-class Gaussian process varies depending on the number of concept values ($v_i$). 
On \SHAPES, it takes approximately 9 minutes using an A100 GPU. 
For \CUB and \CelebA, the training time is longer (due to higher number of concepts) with an average total duration of 45 minutes. The fastest is \DERMA, with a run time of only 2 minutes on average per iteration step.
\change{All reported times assume sequential training over concepts; since each concept GP is trained independently, parallelizing across available workers is possible and can reduce wall-clock time proportionally to the degree of parallelism. We leave this optimization to future work, as sequential training was sufficient in our experimental setting.}

\section{Metrics}
\label{sec:metrics-appx}

We measure \FC as the average of macro $F_1$-score for each concept activation $c_i$ of the models, with $i \in [v] $. 
For \method, measurements of \FC are performed on the concept predictions through $ \{p(c_i \mid \vz)\}$.

We evaluate \ROCAUC for each  component $c_i$, with $i \in [v]$.
For \method, measurements of \ROCAUC are performed on the expected mean of the concept logits of $c_i$, namely $\bbE_{ \vs_i (\vz) \sim \calG \calP_i }(\vA^\top \vs_i(\vz))$.

These metrics allows us to uniformly evaluate both competitors that consider categorical concepts, like \method, and others which model them as their one-hot binarized version, like \LABO, \LFCBM, and \VLGCBM.

We use the disentanglement metric from DCI \citep{eastwood2018framework}, which quantifies the degree to which each learned concept score encodes the variations of a single ground-truth concept, giving a lower score the more single concept scores encode of several ground-truth concepts.

For all VLM-CBM and \CBM{@},
disentanglement is directly measured from by learning the map from concept activations to ground-truth concept values, while for \method we use the expected mean of concept $c_i$ from the GP, \ie $\bbE_{ \vs_i(\vz) \sim \calG \calP_i }(\vA^\top \vs_i(\vz))$.

DCI computes disentanglement as follows. For the general case of
$k$ ground-truth concepts $c_1, \ldots, c_n$ and $v$ learned concepts $\hat c_1, \ldots, \hat c_v$,  the disentanglement score is defined as
\begin{equation}
    D = \sum_{i=1}^{v} \rho_i D_i,
    \quad \text{where} \quad
    \rho_i =
    \frac{\sum_{j=1}^{n} R_{ij}}
         {\sum_{\ell=1}^{k} \sum_{j=1}^{n} R_{\ell j}}.
\end{equation}
This essentially captures the weighted average of the per-learned concept disentanglement scores $D_i$. 
Although $k$ and $v$ may differ in general, in our experiments we always have $v=k$ (specifically, $v=39$ for \CelebA, $v=42$ for \SHAPES, $v=7$ for \DERMA, and $v=112$ for \CUB).
The relative importance matrix $R$ is obtained from a regressor trained to predict each ground-truth concept $c_j$ from the learned concepts $\hat c_i$. 
Each entry $R_{ij}$ quantifies the importance of $\hat c_i$ in predicting $c_j$.

As in the original implementation \citep{eastwood2018framework}, we use a random forest regressor to obtain the coefficients $R_{ij}$, computed as the fraction of decision tree splits on $\hat c_i$ when predicting $c_j$ \citep{breiman1984classification}.

\section{Architectural Details and Hyperparameters}

Here, we report additional details on the different models we tested.  For all experiments, we selected hyperparameters -- including the initial learning rate, initialization values of the GP hyperparameters -- to optimize for concept $F$-score on the validation set.  The batch size was fixed to $512$ for all models.

The weights for all pre-trained models were obtained from \href{https://pytorch.org/vision/main/models.html}{PyTorch} and \href{https://github.com/osmr/imgclsmob}{imgclsmob}.
All methods were trained with sequential training, cf. \cref{sec:preliminaries} (and \citep{koh2020concept}).

\textbf{Concept-bottleneck Models.}  \CBM is implemented using a ResNet18 backbone \citep{he2016deep} followed by a two-stage prediction pipeline: concept prediction and class prediction.
All input images of size 224×224×3 are passed through a ResNet18 \citep{he2016deep} pretrained on ImageNet, which outputs a 1000-dimensional feature vector. 
For \CUB only, we use a ResNet18 finetuned on this dataset (as the original implementation of \citep{koh2020concept}) with output dimension is 200 (matching the number of classes). Also \LFCBM and \LABO adopt this backbone.

The 1000-dimensional (or 200 for \CUB) feature vector is passed through a linear layer projecting to the concept bottleneck. 
This layer returns concept activations, which is trained using standard cross-entropy loss as in \cref{eq:cbm-joint-ce}. 
When datasets are imbalanced, we use a reweighting term to mitigate class imbalances.
The sigmoid activation function is finally applied to the concept scores to model concept probabilities. 
We use Adam optimizer \citep{kingma2014adam}, with early stopping applied when validation loss does not decrease for $8$ consecutive epochs.

The concepts scores are directly fed into the class linear layer mapping from concepts to task labels, giving the class logits.
We use the GLM-SAGA solver to train the linear layer, which optimizes for the elastic-net objective. This steers the training to producing a sparser linear classifier.

\textbf{Label-free CBMs.}  \LFCBM adopt the same backbones of \CBM.
We follow the same setup of the original paper \LFCBM \citep{yang2023language}.
Concept scores are learned by regressing on \CLIP scores by maximizing the cosine similarity.
The linear layer of the classifier is also trained with the GLM-SAGA solver.

\textbf{Language-in-a-Bottle.}  \LABO only features a linear layer to map \CLIP scores to final task labels, which is trained to minimize the cross-entropy loss on ground-truth task labels (as in \citep{yang2023language}) and constraints the columns of the weights to sum to one. 
To prevent using concept filtering steps of the original implementation,
we set \verb|clip cutoff=0.0| and \verb|interpretability cutoff=0.0|). This forces \LABO to use all concept inside the ground-truth vocabulary.

\textbf{Vision-Language-guided CBMs.}  Following \citep{srivastava2024vlgcbm}, we implement \VLGCBM with the \CLIP backbone. Concept annotation is obtained  prompting Grounding Dino \citep{liu2024grounding} (while setting \verb|crop to concept| and \verb|confidence threshold=0.15|) passing both input and label pairs. 
The concept extractor is successively trained by fitting an MLP on \CLIP (setting \verb|cbl hidden layers=3| and \verb|cbl lr=0.001|).
The final linear layer is also trained GLM-SAGA.

\textbf{Vision-and-HUman-guided CBMs (\method{s}).}  We first process the inputs (\ie images) using \CLIP and gather the embeddings to feed into the GPs.  Then, our method uses the trained GPs as a backbone for extracting concept activations. 

All GPs are trained using a scheduler for the learning rate, \texttt{ReduceLROnPlateau} with patience $80$ and an exponential decay of $0.8$. The learning rate starts at $0.01$ and the training lasts for $8$k epochs or when the loss gets lower than $10^{-7}$.

The classifier consists of a linear layer trained with cross entropy loss on annotated task labels with a learning rate of $0.001$ and stopping criterion based on whether the validation stalls for 3 epochs.

\end{document}